\documentclass{article}

\usepackage[utf8]{inputenc}
\usepackage[T1]{fontenc}
\usepackage{lineno}
\usepackage{tipa}

\usepackage{url}
\usepackage{doi}
\usepackage[normalem]{ulem}  
\usepackage{booktabs}       
\usepackage{longtable}
\usepackage[draft]{minted}  
\usepackage{gensymb}
\usepackage[detect-all]{siunitx}
\usepackage{placeins}
\usepackage{float}
\usepackage{multirow}

\usepackage[parfill]{parskip}

\usepackage[%
    a4paper,
    inner=25mm,
    outer=25mm,
    top=25mm,
    bottom=25mm,
    marginparsep=5mm,
    marginparwidth=40mm,
]{geometry}

\newlength{\defbaselineskip}
\usepackage[tt=false, type1=true]{libertine}
\usepackage[varqu]{zi4}
\usepackage[libertine]{newtxmath}

\usepackage{fancyhdr}
\usepackage{hyperref}       
\usepackage{url}            
\usepackage[dvipsnames,table]{xcolor} 
\usepackage{graphicx}       
\usepackage{mathtools}
\usepackage{amsmath} 
\usepackage{amsfonts}       
\usepackage{nicefrac}       
\usepackage{microtype}      
\usepackage{array}
\usepackage{times}

\usepackage{authblk} 

\makeatletter
\renewcommand\@author{
    \AB@authlist\\[\affilsep]
    \raggedright
    \parindent-0.7em
    \AB@affillist
}
\makeatother

\newcommand{\appref}[1]{\hyperref[#1]{Appendix~\ref*{#1}}}

\usepackage{bbding}
\definecolor{cold}{HTML}{008acd}
\newcommand{\frozen}{\raisebox{-.15ex}{\color{cold}\SnowflakeChevron}}
\DeclareUnicodeCharacter{2744}{\frozen}

\def\unfrozen{\raisebox{-.15ex}{\includegraphics[height=1em]{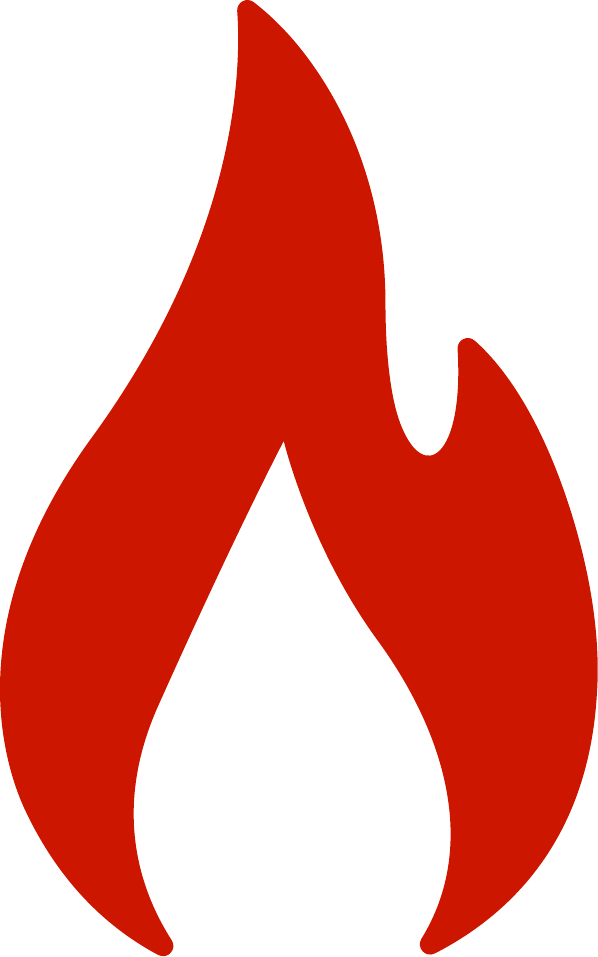}}}
\DeclareUnicodeCharacter{1F525}{\unfrozen}

\usepackage{natbib}

\newcommand{\hide}[1]{}
\newcommand{\hider}[1]{}
\newcommand{\note}[1]{}
\newcommand{\todo}[1]{}

\newcommand{\Rshd}{\rotatebox[origin=c]{180}{$\Lsh$}}
\DeclareUnicodeCharacter{21B3}{\Rshd}

\DeclareUnicodeCharacter{2194}{$\leftrightarrow$}

\newcommand{\terr}[1]{{\footnotesize{\ensuremath{{\color{gray}\,\pm{}\,#1}}}}}
\newcommand{\noval}[0]{\multicolumn{1}{c}{--}}

\usepackage{pifont}

\DeclareRobustCommand{\bigO}{%
  \text{\usefont{OMS}{cmsy}{m}{n}O}%
}

\newcommand{\aref}[1]{\appref{#1}}

\definecolor{linkcolor}{HTML}{991408}  
\definecolor{citecolor}{HTML}{2E7E2A}  
\definecolor{filecolor}{HTML}{131877}  
\definecolor{menucolor}{HTML}{727500}  
\definecolor{runcolor} {HTML}{137776}  
\definecolor{urlcolor} {HTML}{0a2bbf}  
\hypersetup{%
    colorlinks=true,%
    linkcolor=linkcolor,%
    citecolor=citecolor,%
    filecolor=filecolor,%
    menucolor=menucolor,%
    runcolor=runcolor,%
    urlcolor=urlcolor%
}

\newcommand{\genus}[1]{\textit{#1}}
\newcommand{\maybefootnote}[1]{ (#1)}
\renewcommand{\maybefootnote}[1]{\footnote{#1}}

\newcommand{\arxivonly}[1]{}  
\renewcommand{\arxivonly}[1]{#1}  

\makeatletter
\def\PYG@reset{\let\PYG@it=\relax \let\PYG@bf=\relax%
    \let\PYG@ul=\relax \let\PYG@tc=\relax%
    \let\PYG@bc=\relax \let\PYG@ff=\relax}
\def\PYG@tok#1{\csname PYG@tok@#1\endcsname}
\def\PYG@toks#1+{\ifx\relax#1\empty\else%
    \PYG@tok{#1}\expandafter\PYG@toks\fi}
\def\PYG@do#1{\PYG@bc{\PYG@tc{\PYG@ul{%
    \PYG@it{\PYG@bf{\PYG@ff{#1}}}}}}}
\def\PYG#1#2{\PYG@reset\PYG@toks#1+\relax+\PYG@do{#2}}

\@namedef{PYG@tok@w}{\def\PYG@tc##1{\textcolor[rgb]{0.73,0.73,0.73}{##1}}}
\@namedef{PYG@tok@c}{\let\PYG@it=\textit\def\PYG@tc##1{\textcolor[rgb]{0.24,0.48,0.48}{##1}}}
\@namedef{PYG@tok@cp}{\def\PYG@tc##1{\textcolor[rgb]{0.61,0.40,0.00}{##1}}}
\@namedef{PYG@tok@k}{\let\PYG@bf=\textbf\def\PYG@tc##1{\textcolor[rgb]{0.00,0.50,0.00}{##1}}}
\@namedef{PYG@tok@kp}{\def\PYG@tc##1{\textcolor[rgb]{0.00,0.50,0.00}{##1}}}
\@namedef{PYG@tok@kt}{\def\PYG@tc##1{\textcolor[rgb]{0.69,0.00,0.25}{##1}}}
\@namedef{PYG@tok@o}{\def\PYG@tc##1{\textcolor[rgb]{0.40,0.40,0.40}{##1}}}
\@namedef{PYG@tok@ow}{\let\PYG@bf=\textbf\def\PYG@tc##1{\textcolor[rgb]{0.67,0.13,1.00}{##1}}}
\@namedef{PYG@tok@nb}{\def\PYG@tc##1{\textcolor[rgb]{0.00,0.50,0.00}{##1}}}
\@namedef{PYG@tok@nf}{\def\PYG@tc##1{\textcolor[rgb]{0.00,0.00,1.00}{##1}}}
\@namedef{PYG@tok@nc}{\let\PYG@bf=\textbf\def\PYG@tc##1{\textcolor[rgb]{0.00,0.00,1.00}{##1}}}
\@namedef{PYG@tok@nn}{\let\PYG@bf=\textbf\def\PYG@tc##1{\textcolor[rgb]{0.00,0.00,1.00}{##1}}}
\@namedef{PYG@tok@ne}{\let\PYG@bf=\textbf\def\PYG@tc##1{\textcolor[rgb]{0.80,0.25,0.22}{##1}}}
\@namedef{PYG@tok@nv}{\def\PYG@tc##1{\textcolor[rgb]{0.10,0.09,0.49}{##1}}}
\@namedef{PYG@tok@no}{\def\PYG@tc##1{\textcolor[rgb]{0.53,0.00,0.00}{##1}}}
\@namedef{PYG@tok@nl}{\def\PYG@tc##1{\textcolor[rgb]{0.46,0.46,0.00}{##1}}}
\@namedef{PYG@tok@ni}{\let\PYG@bf=\textbf\def\PYG@tc##1{\textcolor[rgb]{0.44,0.44,0.44}{##1}}}
\@namedef{PYG@tok@na}{\def\PYG@tc##1{\textcolor[rgb]{0.41,0.47,0.13}{##1}}}
\@namedef{PYG@tok@nt}{\let\PYG@bf=\textbf\def\PYG@tc##1{\textcolor[rgb]{0.00,0.50,0.00}{##1}}}
\@namedef{PYG@tok@nd}{\def\PYG@tc##1{\textcolor[rgb]{0.67,0.13,1.00}{##1}}}
\@namedef{PYG@tok@s}{\def\PYG@tc##1{\textcolor[rgb]{0.73,0.13,0.13}{##1}}}
\@namedef{PYG@tok@sd}{\let\PYG@it=\textit\def\PYG@tc##1{\textcolor[rgb]{0.73,0.13,0.13}{##1}}}
\@namedef{PYG@tok@si}{\let\PYG@bf=\textbf\def\PYG@tc##1{\textcolor[rgb]{0.64,0.35,0.47}{##1}}}
\@namedef{PYG@tok@se}{\let\PYG@bf=\textbf\def\PYG@tc##1{\textcolor[rgb]{0.67,0.36,0.12}{##1}}}
\@namedef{PYG@tok@sr}{\def\PYG@tc##1{\textcolor[rgb]{0.64,0.35,0.47}{##1}}}
\@namedef{PYG@tok@ss}{\def\PYG@tc##1{\textcolor[rgb]{0.10,0.09,0.49}{##1}}}
\@namedef{PYG@tok@sx}{\def\PYG@tc##1{\textcolor[rgb]{0.00,0.50,0.00}{##1}}}
\@namedef{PYG@tok@m}{\def\PYG@tc##1{\textcolor[rgb]{0.40,0.40,0.40}{##1}}}
\@namedef{PYG@tok@gh}{\let\PYG@bf=\textbf\def\PYG@tc##1{\textcolor[rgb]{0.00,0.00,0.50}{##1}}}
\@namedef{PYG@tok@gu}{\let\PYG@bf=\textbf\def\PYG@tc##1{\textcolor[rgb]{0.50,0.00,0.50}{##1}}}
\@namedef{PYG@tok@gd}{\def\PYG@tc##1{\textcolor[rgb]{0.63,0.00,0.00}{##1}}}
\@namedef{PYG@tok@gi}{\def\PYG@tc##1{\textcolor[rgb]{0.00,0.52,0.00}{##1}}}
\@namedef{PYG@tok@gr}{\def\PYG@tc##1{\textcolor[rgb]{0.89,0.00,0.00}{##1}}}
\@namedef{PYG@tok@ge}{\let\PYG@it=\textit}
\@namedef{PYG@tok@gs}{\let\PYG@bf=\textbf}
\@namedef{PYG@tok@gp}{\let\PYG@bf=\textbf\def\PYG@tc##1{\textcolor[rgb]{0.00,0.00,0.50}{##1}}}
\@namedef{PYG@tok@go}{\def\PYG@tc##1{\textcolor[rgb]{0.44,0.44,0.44}{##1}}}
\@namedef{PYG@tok@gt}{\def\PYG@tc##1{\textcolor[rgb]{0.00,0.27,0.87}{##1}}}
\@namedef{PYG@tok@err}{\def\PYG@bc##1{{\setlength{\fboxsep}{\string -\fboxrule}\fcolorbox[rgb]{1.00,0.00,0.00}{1,1,1}{\strut ##1}}}}
\@namedef{PYG@tok@kc}{\let\PYG@bf=\textbf\def\PYG@tc##1{\textcolor[rgb]{0.00,0.50,0.00}{##1}}}
\@namedef{PYG@tok@kd}{\let\PYG@bf=\textbf\def\PYG@tc##1{\textcolor[rgb]{0.00,0.50,0.00}{##1}}}
\@namedef{PYG@tok@kn}{\let\PYG@bf=\textbf\def\PYG@tc##1{\textcolor[rgb]{0.00,0.50,0.00}{##1}}}
\@namedef{PYG@tok@kr}{\let\PYG@bf=\textbf\def\PYG@tc##1{\textcolor[rgb]{0.00,0.50,0.00}{##1}}}
\@namedef{PYG@tok@bp}{\def\PYG@tc##1{\textcolor[rgb]{0.00,0.50,0.00}{##1}}}
\@namedef{PYG@tok@fm}{\def\PYG@tc##1{\textcolor[rgb]{0.00,0.00,1.00}{##1}}}
\@namedef{PYG@tok@vc}{\def\PYG@tc##1{\textcolor[rgb]{0.10,0.09,0.49}{##1}}}
\@namedef{PYG@tok@vg}{\def\PYG@tc##1{\textcolor[rgb]{0.10,0.09,0.49}{##1}}}
\@namedef{PYG@tok@vi}{\def\PYG@tc##1{\textcolor[rgb]{0.10,0.09,0.49}{##1}}}
\@namedef{PYG@tok@vm}{\def\PYG@tc##1{\textcolor[rgb]{0.10,0.09,0.49}{##1}}}
\@namedef{PYG@tok@sa}{\def\PYG@tc##1{\textcolor[rgb]{0.73,0.13,0.13}{##1}}}
\@namedef{PYG@tok@sb}{\def\PYG@tc##1{\textcolor[rgb]{0.73,0.13,0.13}{##1}}}
\@namedef{PYG@tok@sc}{\def\PYG@tc##1{\textcolor[rgb]{0.73,0.13,0.13}{##1}}}
\@namedef{PYG@tok@dl}{\def\PYG@tc##1{\textcolor[rgb]{0.73,0.13,0.13}{##1}}}
\@namedef{PYG@tok@s2}{\def\PYG@tc##1{\textcolor[rgb]{0.73,0.13,0.13}{##1}}}
\@namedef{PYG@tok@sh}{\def\PYG@tc##1{\textcolor[rgb]{0.73,0.13,0.13}{##1}}}
\@namedef{PYG@tok@s1}{\def\PYG@tc##1{\textcolor[rgb]{0.73,0.13,0.13}{##1}}}
\@namedef{PYG@tok@mb}{\def\PYG@tc##1{\textcolor[rgb]{0.40,0.40,0.40}{##1}}}
\@namedef{PYG@tok@mf}{\def\PYG@tc##1{\textcolor[rgb]{0.40,0.40,0.40}{##1}}}
\@namedef{PYG@tok@mh}{\def\PYG@tc##1{\textcolor[rgb]{0.40,0.40,0.40}{##1}}}
\@namedef{PYG@tok@mi}{\def\PYG@tc##1{\textcolor[rgb]{0.40,0.40,0.40}{##1}}}
\@namedef{PYG@tok@il}{\def\PYG@tc##1{\textcolor[rgb]{0.40,0.40,0.40}{##1}}}
\@namedef{PYG@tok@mo}{\def\PYG@tc##1{\textcolor[rgb]{0.40,0.40,0.40}{##1}}}
\@namedef{PYG@tok@ch}{\let\PYG@it=\textit\def\PYG@tc##1{\textcolor[rgb]{0.24,0.48,0.48}{##1}}}
\@namedef{PYG@tok@cm}{\let\PYG@it=\textit\def\PYG@tc##1{\textcolor[rgb]{0.24,0.48,0.48}{##1}}}
\@namedef{PYG@tok@cpf}{\let\PYG@it=\textit\def\PYG@tc##1{\textcolor[rgb]{0.24,0.48,0.48}{##1}}}
\@namedef{PYG@tok@c1}{\let\PYG@it=\textit\def\PYG@tc##1{\textcolor[rgb]{0.24,0.48,0.48}{##1}}}
\@namedef{PYG@tok@cs}{\let\PYG@it=\textit\def\PYG@tc##1{\textcolor[rgb]{0.24,0.48,0.48}{##1}}}

\def\PYGZus{\char`\_}
\def\PYGZob{\char`\{}
\def\PYGZcb{\char`\}}

\def\PYGZdq{\char`\"}

\makeatother

\title{BenthicNet: A global compilation of seafloor images for deep learning applications}
\date{}

\author[1,$\dag$]{Scott~C.~Lowe}
\author[2,3,$\dag$,*]{Benjamin~Misiuk}
\author[4,$\dag$]{Isaac~Xu}
\author[4]{\\Shakhboz~Abdulazizov}
\author[5,6]{Amit~R.~Baroi}
\author[7]{Alex~C.~Bastos}
\author[8]{Merlin~Best}
\author[9]{Vicki~Ferrini}
\author[10,11]{Ariell~Friedman}
\author[12]{Deborah~Hart}
\author[13]{Ove~Hoegh-Guldberg}
\author[14]{Daniel~Ierodiaconou}
\author[15]{Julia~Mackin-McLaughlin}
\author[13]{Kathryn~Markey}
\author[7]{Pedro~S.~Menandro}
\author[16]{Jacquomo~Monk}
\author[17]{Shreya~Nemani}
\author[18]{John~O'Brien}
\author[16]{Elizabeth~Oh}
\author[19]{Luba~Y.~Reshitnyk}
\author[17]{Katleen~Robert}
\author[13]{Chris~M.~Roelfsema}
\author[18]{Jessica~A.~Sameoto}
\author[20]{Alexandre~C.~G.~Schimel}
\author[21]{Jordan~A.~Thomson}
\author[18]{Brittany~R.~Wilson}
\author[18]{Melisa~C.~Wong}
\author[22,$\ddag$]{Craig~J.~Brown}
\author[4,$\ddag$]{Thomas~Trappenberg}

\affil[1]{Vector Institute, Toronto, Ontario, Canada}
\affil[2]{Memorial University of Newfoundland, Department of Geography, St. John's, Newfoundland, Canada}
\affil[3]{Memorial University of Newfoundland, Department of Earth Sciences, St. John's, Newfoundland, Canada}
\affil[4]{Dalhousie University, Faculty of Computer Science, Halifax, Nova Scotia, Canada}
\affil[5]{Dalhousie University, School for Resource and Environmental Studies, Halifax, Nova Scotia, Canada}
\affil[6]{Dalhousie University, DeepSense, Halifax, Nova Scotia, Canada}
\affil[7]{Universidade Federal do Espírito Santo, Departamento de Oceanografia e Ecologia, Vitória, Brazil}
\affil[8]{Fisheries and Oceans Canada, Marine Spatial Ecology and Analysis Section, Institute of Ocean Sciences, Sidney, British Columbia, Canada}
\affil[9]{Columbia University, Lamont-Doherty Earth Observatory, Palisades, New York, USA}
\affil[10]{University of Sydney, Australian Centre for Field Robotics, Sydney, New South Wales, Australia}
\affil[11]{Greybits Engineering, Sydney, New South Wales, Australia}
\affil[12]{National Oceanic and Atmospheric Administration Northeast Fisheries Science Center, Woods Hole, Massachusetts, USA}
\affil[13]{University of Queensland, School of the Environment, Brisbane, Queensland, Australia}
\affil[14]{Deakin University, School of Life and Environmental Sciences, Warrnambool, Victoria, Australia}
\affil[15]{Oxy Occidental College, Vantuna Research Group, Los Angeles, California, USA}
\affil[16]{University of Tasmania, Institute for Marine and Antarctic Studies, Hobart, Tasmania, Australia}
\affil[17]{Fisheries and Marine Institute of Memorial University of Newfoundland, School of Ocean Technology, St.~John's, Newfoundland, Canada}
\affil[18]{Fisheries and Oceans Canada, Bedford Institute of Oceanography, Dartmouth, Nova Scotia, Canada}
\affil[19]{Hakai Institute, Heriot Bay, British Columbia, Canada}
\affil[20]{Geological Survey of Norway (NGU), Trondheim, Norway}
\affil[21]{Ecology Action Centre, Halifax, Nova Scotia, Canada}
\affil[22]{Dalhousie University, Department of Oceanography, Halifax, Nova Scotia, Canada\vspace{0.5em}}
\affil[*]{Corresponding author: Benjamin~Misiuk \texttt{<bmisiuk@mun.ca>}}
\affil[$\dag$]{Joint first author}
\affil[$\ddag$]{Joint last author}

\begin{document}
\newgeometry{top=10mm, bottom=15mm}
\maketitle
\thispagestyle{empty}
\begin{abstract}
Advances in underwater imaging enable collection of extensive seafloor image datasets necessary for monitoring important benthic ecosystems. The ability to collect seafloor imagery has outpaced our capacity to analyze it, hindering mobilization of this crucial environmental information. Machine learning approaches provide opportunities to increase the efficiency with which seafloor imagery is analyzed, yet large and consistent datasets to support development of such approaches are scarce. Here we present BenthicNet: a global compilation of seafloor imagery designed to support the training and evaluation of large-scale image recognition models. An initial set of over 11.4 million images was collected and curated to represent a diversity of seafloor environments using a representative subset of 1.3 million images. These are accompanied by 3.1 million annotations translated to the CATAMI scheme, which span \num{190000} of the images. A large deep learning model was trained on this compilation and preliminary results suggest it has utility for automating large and small-scale image analysis tasks. The compilation and model are made openly available for reuse at \doi{10.20383/103.0614}.
\end{abstract}
\clearpage
\restoregeometry

\section{Background \& Summary}

Spatial data products convey information that is necessary to achieve marine management goals \citep{harris_why_2020}, including monitoring species or habitats of interest, informing policy decisions, and guiding sustainable ocean resource use \citep{baker_habitat_2020}. The creation of seafloor spatial data products is broadly referred to as ``benthic habitat mapping'' \citep{brown_benthic_2011, misiuk_brown_2023}, which describes both biotic and abiotic mapping elements. High quality spatial data underpins accurate benthic habitat maps, and advances in marine sampling technologies and techniques have increased capacity to collect and analyze benthic data effectively.

Underwater imagery, including both still photographs and video, is among the most common forms of data used to inform benthic habitat mapping. Seabed imagery has great utility for characterizing benthic environments for several reasons: it is non-invasive and minimally destructive, it may be collected remotely, it may be analyzed for multiple purposes (e.g. biology, geology), and it is more efficient to collect and store than physical samples (e.g. grabs, cores, preserved specimen). In addition to manual \textit{in situ} (e.g. snorkeling, SCUBA) or surface (e.g. drop camera) deployment, imagery is increasingly collected using automated and remote underwater vehicles (AUVs, ROVs). Each of these deployment methods, but particularly AUVs and ROVs, have the potential to generate large volumes of imagery data \citep{williams2010auv, gonzalez2014catlin}. In addition to large data volumes, image datasets also often have characteristics of spatial, and therefore environmental, redundancy \citep{foster_choosing_2014} --- for example, where proximal image frames extracted from video data depict similar biological or geological attributes due to positive spatial autocorrelation \citep{kendall2005benthic, misiuk2019mapping}. 

Benthic image data is traditionally analyzed by a trained operator, yet this is often inefficient given the volume and spatial redundancy that typify benthic image datasets. Depending on the detail of analysis (e.g. level of taxonomic identification), there is now capacity to collect image data faster than it can be analyzed \citep{schoening2016} --- particularly in the case of automated platforms such as AUVs. The manual classification, annotation, and labelling of seabed imagery therefore acts as a bottleneck in the habitat mapping workflow \citep{coralnet}, and it is common to analyze only small portions of large image datasets to expedite the production of spatial data products. These inefficiencies offer opportunities for automation.

Machine learning can facilitate the automation of manual and subjective tasks. A machine learning model is created by collecting many samples of example data for a given task, then training a model that accurately maps input samples to the desired target outputs. In particular, deep learning allows us to build the complex models necessary to solve challenging tasks that would be laborious for a human to perform \citep{deeplearning2015,Schmidhuber2015,Goodfellow-book}.
Deep learning models have revolutionized computer vision over the last 10 years \citep{alexnet,resnet,turinglecture2021}, and have been successfully applied to a variety of image processing tasks such as classification \citep{alexnet, resnet, vgg-19, inception-v3, efficientnet, Dosovitskiy2020}, semantic and instance segmentation \citep{unet, yolo, deep-segmentation-survey}, and image generation \citep{GAN, GLIDE, DALLE2, latent_diffusion}.
Some models have even attained human-level or superhuman performance at narrow tasks \citep{imagenet,resnet,Santoro2016}.
The dominant network architectures used in the field of computer vision are currently convolutional neural networks \citep{LeNet}, yet other attention-based architectures such as vision transformers \citep{Dosovitskiy2020} are increasingly applied.

Successfully training large-scale deep learning models from scratch requires large volumes of data. However, a deep neural network that has previously been trained on one task can be repurposed for a new task, provided the new task uses similar input stimuli to that used when training the original network.
This process, known as transfer learning, can save considerable resources, since retraining or ``fine-tuning'' a model requires much fewer computational resources than training a whole new one. This can enable the learning of novel tasks from labelled datasets that would otherwise be too small to support training a deep network from scratch. Transfer learning is possible because the early layers of a deep neural network need to learn to see image stimuli first in order to comprehend and process them. The subtask of seeing and understanding the image stimuli constitutes most of the complexity of any image processing task, and this subtask is common to all tasks involving imagery from that domain. For image data of the natural world, transfer learning is typically performed by reusing models pretrained on the widely available ILSVRC-2012 (ImageNet-1k) dataset \citep{imagenet}, consisting of 1.28M photographs of real world objects scraped from image hosting websites.
However, this dataset comprises terrestrial and anthropocentric objects and scenes, and does not represent subaqueous  environments. The difference in the domain of the input data may limit the capacity for transfer learning.

Development of  large-scale models using compilations of benthic imagery that are suitable for transfer learning purposes would be ideal, yet this is made difficult by a lack of universal labels for seabed features. One of the primary difficulties associated with developing deep learning models in this context is that, unlike terrestrial and anthropocentric images, there is no objective label for many seabed habitats, biological communities, substrate types, or organisms. Indeed, a number of different classification schemes are used to label benthic features \citep{catami, NISB2007, cmecs}. Because no single vocabulary is universally applied to describe these features, we currently lack large sets of consistently labelled images that are necessary for training deep learning models for benthic environments. We note an outstanding need to develop standardized protocols for the translation of common marine image labelling schemes. 

Self-supervised learning (SSL) is a recent technique in which models can learn to understand their stimuli without the use of manually annotated data \citep{SSLCookbook, SimCLR, MoCo-v1, chenbigssl, BYOL, SimSiam, MoCo-v3, DINO, masked_autoencoder}.  Instead of using labelled data, self-supervised models learn to solve a pretext task that can be automatically constructed from the input data itself. SSL enables the training of large-scale models on unlabelled imagery, which can be collected at scale more easily than annotated imagery. Models trained with SSL have already learnt to see and understand the stimuli of interest, and can subsequently be used for transfer learning onto specific tasks, even if there is only a limited amount of annotated data available for the new task.

SSL may enable the training of deep learning models on large-scale benthic image datasets for the purposes of transfer learning on smaller novel tasks (e.g. site-specific habitat labelling), despite the lack of large consistently labelled image datasets. Cumulatively, adequate volumes of benthic image data currently exist to support the development of SSL models, but they are spread globally among various research groups, government data portals, and open data repositories. There is a need to compile and curate datasets for the development of large-scale image recognition models. Such compilations must include images from a range of biomes, depths, and physical oceanographic conditions in order to adequately represent the global heterogeneity of benthic environments. Additionally, data should be included from an array of acquisition platforms and camera configurations to represent the variability in image characteristics (e.g. lighting, resolution, quality, perspective) that arise from non-standardized image data collection methods.

The intended applications and scope for a benthic habitat machine learning image dataset dictate qualities that images should possess to be useful for automating tasks in this context. Unlike imagery that is focused solely on specific biota, benthic habitat images often depict a broader area (e.g. on the order of m${^2}$), which necessarily includes the seafloor. The goal of analyzing such data is often to broadly categorize the benthic environment, potentially including both biotic and abiotic elements \citep{ierodiaconou2018combining}. Biotic characterization may include descriptions of individual organisms \citep{BAUMSTARK201683} or community composition \citep{buhl2020classification}, while abiotic components include description of substrate, sediment bed forms, heterogeneity, rugosity, and relief \citep{espinosa2015assessment, neves2014mapping, wienberg2013coral}. For these reasons, benthic habitat information is often summarized at the whole-image level --- for example, by assigning one or several ``habitat'' labels to an entire image using a pre-defined scheme \citep{NISB2007, cmecs, GALPARSORO20122630}, or by aggregating individual labels indicating presence or absence, abundance, or percentage cover of individual habitat components, which may be labelled using a more detailed vocabulary \citep{catami}. It is therefore useful for benthic habitat images to depict a broad enough area so that both abiotic and biotic habitat components may be recognized. 

The whole-image labels that typify benthic habitat image datasets may differ from other forms of marine image labelling that focus on locating specific objects, semantic labelling, bounding boxes, and masking. These forms of labelling are well suited to applications focusing on single taxa, pelagic biota, and object detection or tracking, and efforts to establish extensive image datasets for those applications are also underway. FathomNet \citep{boulais2020fathomnet, katija_fathomnet}, for example, contains over \num{100000} images and over \num{200000} localization labels (i.e. bounding boxes) focused generally on marine biota, while \citet{orenstein2015whoi} present a dataset of \num{3.4} million plankton images that have been labelled and used to train deep learning models. \Citet{trashcan} have established the TrashCan dataset, containing over \num{7000} images of marine debris with corresponding bounding box and segmentation masks, which has been used to develop object detection and semantic segmentation models. Other comparable image datasets include WildFish for classifying fish species \citep{wildfish}, the OUC-vision large-scale underwater image database for underwater salient object detection \citep{ouc}, and the Brackish dataset \citep{pedersen2019detection} for detecting fish, crabs and starfish in brackish waters. Multiple datasets have been established to support the automated annotation of coral imagery including Moorea Labeled Corals \citep{beijbom2012automated}, the Gulf of Eilat dataset \citep{raphael2020deep}, and notably, CoralNet \citep{coralnet, coralnet2021}. In addition to serving as a data repository where users can upload and share underwater image data and point labels, CoralNet provides a web interface to facilitate labelling and development of image recognition models. Several other comparable data portals and software packages enable the labelling and centralization of marine image data in this way (e.g. FathomNet, \citealp{katija_fathomnet}; SQUIDLE+; BIIGLE, \citealp{biigle}; VIAME).

Here we describe BenthicNet: a global compilation of seafloor images that is designed to support development of automated image processing tools for benthic habitat data. With this compilation, we strive to obtain thematic diversity by (i) compiling benthic habitat images from locations around the world, and (ii) representing habitats from a broad range of marine environments. The compiled dataset is assessed for these qualities. Additionally, we aim to achieve diversity of non-thematic image characteristics (e.g. image quality, lighting, perspective) by obtaining data from a range of acquisition platforms and camera configurations. The dataset is presented in three parts: a diverse collection of over 11 million seafloor images from around the world, provided without labels (BenthicNet-11M); a rarefied subset of 1.3 million images, selected to maintain diversity in the imagery while reducing redundancy and volume (BenthicNet-1M); and a collection of \num{188688} labelled images bearing 3.1 million annotations (BenthicNet-Labelled). We provide a large SSL model pretrained on BenthicNet-1M, and demonstrate its application using examples from BenthicNet-Labelled. The  compilation and SSL model are made openly available to foster further development and assessment of benthic image automation tools.

\section{Methods}

In order to achieve a diverse collection of benthic habitat images for training deep learning models, data spanning a range of environments and geographies were obtained from a variety of sources. These initially included project partners and research contacts, which were leveraged to establish additional data partnerships with individuals, academic and not-for-profit research groups, and government organizations. The largest data volumes were eventually obtained from several academic, government, and third-party public data repositories. The acquisition of labelled data was prioritized in all cases, but extensive high quality unlabelled data collections were also included where feasible. The desired format for each dataset was a single folder containing unique images, accompanied by a single comma separated value (CSV) file indicating, at a minimum, the dataset, file name, latitude, longitude, date and time of acquisition, URL (if hosted online) and label(s) (if provided) for each image.

\subsection{Data compilation and quality control}
\label{s:Data-compilation-and-quality-control}

Labelled benthic image data was initially obtained from project collaborators, data partners, and opportunistic sources such as academic journal supplementary materials. The formats and varieties of data were diverse, including collections of images with spreadsheet metadata, images with metadata contained in file names, GIS files containing images from which metadata was extracted, lists of URL image links, and raw video with text file annotations. Datasets that were not formatted as a single folder of images or list of URL links with CSV metadata were re-formatted upon receipt. Metadata contained in image file names was parsed and used to construct a metadata CSV file where necessary. Image data contained within GIS files was extracted using ArcGIS Pro and the ArcPy Python package, along with geographic information and other metadata contained within the files. All geographic coordinates were converted to decimal degrees using the WGS~84 datum. Data obtained as video files were subsampled by extracting still frames according to their metadata using FFmpeg. After formatting, all datasets were subjected to quality control checks for missing entries, duplicates, label consistency, image quality, and matches between images and metadata. Quality control of image labels was performed by sampling the metadata and comparing labels to corresponding images for each dataset or data source, but this was not exhaustive. Datasets where notable label inconsistencies were detected were rejected. Data columns were renamed to match a standardized format for the BenthicNet dataset. All quality control and formatting was completed using R and Python. The dataset sources are summarized in \autoref{tab:datasets}. Additional detail on the individual datasets is provided with the BenthicNet metadata \citep{BenthicNet_FRDR}.

\subsubsection{Individual contributions}

A number of datasets were contributed by individual project partners; several of these were from eastern Canada. The Seascape Ecology and Mapping (SEAM) Lab at Dalhousie University provided three datasets for the BenthicNet compilation from this region. Still images were provided ($n=2281$) that were extracted from passive drop down video drifts conducted in the Bay of Fundy at 281 sites between 2017--2019 using a 4k camera system \citep{wilson2021}. Whole-image labels were supplied according to site-specific ``benthoscapes'' interpreted by the image analyst, which are recognizable combinations of dominant substrate type and biological characteristics \citep{brown_benthic_2011, misiuk_brown_2023}. All megafauna were additionally identified to the highest possible taxonomic resolution for each image. A dataset of high definition benthic photographs ($n=4064$) was also provided from surveys conducted between 2009--2014 at the St Anns Bank marine protected area \citep{lacharite2018}, which included whole-image benthoscape labels defined for the site. Finally, the SEAM lab contributed photographs of the seabed ($n=62$) used for the 2017 R2Sonic Multispectral Challenge in the Bedford Basin, Nova Scotia \citep{brown2019}, which included broad whole-image substrate descriptions and, occasionally, biological observations. The 4D Oceans lab at the Fisheries and Marine Institute of Memorial University of Newfoundland provided still images ($n=3000$) extracted from underwater video, as part of the project ``Coastal Habitat Mapping of Placentia Bay'' conducted off the coast of Newfoundland, which included whole-image substrate-derived bottom class labels \citep{mackin_2022, nemani_2022}. The Ecology Action Centre (\href{https://ecologyaction.ca/}{EAC}) provided \num{1220} images collected by citizen scientists via Go Pro-mounted kayak between 2019--2021 at shallow eelgrass sites in Nova Scotia. These included whole-image labels for the presence or absence of eelgrass (\textit{Zostera marina}). 

Several datasets collected by researchers at Memorial University of Newfoundland (MUN) were also contributed from northern Canada. These included \num{895} images collected for a benthic mapping project in Frobisher Bay, Nunavut, between 2015--2016 \citep{misiuk2019frobisher}; 1059 images from Wager Bay, Nunavut, collected in collaboration with Parks Canada as part of the Ukkusiksalik National Park Marine Baseline Data Collection Project; 541 images from Chesterfield Inlet, Nunavut, collected for a local benthic habitat mapping project conducted in coordination with the Government of Nunavut, and University of Manitoba; and \num{8443} images from the area around Qikiqtarjuaq, Nunavut, which were obtained as part of a mapping campaign to monitor a locally harvested soft-shell clam population \citep{misiuk2019mapping}. These datasets were each accompanied by site-specific whole-image labels describing the dominant substrate types visible in each image. 

Several image datasets were provided by the Hakai Institute from western Canada. A total of \num{8787} images were obtained from nearshore benthic surveys conducted between 2017--2020 from sites on the central coast of British Columbia and sites within Pacific Rim National Park Reserve (PRNPR). This data was comprised of still images from ROV deployments and GIS-annotated drop camera videos collected primarily for the purposes of mapping eelgrass meadows (\textit{Zostera marina}). Still images were extracted from videos using the methods described above (i.e. using FFmpeg). Whole-image labels were provided corresponding to the dominant visible substrate and vegetation type present in each image.

Individual datasets were also acquired from outside Canada. The Marine Geosciences Lab (LaboGeo) at Universidade Federal do Espirito Santo (UFES) provided quadrat sample images acquired by drop camera during rhodolith surveys off the east coast of Brazil between 2015--2020 \citep{menandro_2022, menandro_2023}. These were cropped to remove the quadrat frame, and \num{360} images were included in the BenthicNet compilation. Whole-image labels were provided that identify the presence of rhodoliths and select biogenic substrate types. A dataset of \num{505} images was provided by the Hierarchical Anticipatory Learning (HAL) lab at Dalhousie University, which was collected from Ocho Rios, Jamaica, in shallow water by snorkeler in 2022. Images were unlabelled, and comprised coral reef and a range of substrate types.

\subsubsection{DFO}

Fisheries and Oceans Canada (\href{https://www.dfo-mpo.gc.ca/}{DFO}) is a federal institution responsible for managing many of Canada's marine resources. DFO provided three separate contributions to the BenthicNet compilation. The Population Ecology Division at the Bedford Institute of Oceanography (BIO) contributed 645 annotated images from George's Bank, which separates the Gulf of Maine from the Northwest Atlantic. These images were collected by the Geological Survey of Canada (\href{https://www.nrcan.gc.ca/science-and-data/research-centres-and-labs/geological-survey-canada/17100}{GSC}) Atlantic for programs under Natural Resources Canada (NRCan) using the Campod digitial camera system deployed from the CCGS \textit{Hudson} in 2000 \citep{george2000} and 2002 \citep{george2002}. Annotations included whole-image benthoscape labels describing the primary substrate and presence of characteristic biota. Benthic images were also contributed from a GSC survey on German Bank off the southwest coast of Nova Scotia in 2003 \citep{german2003} using Campod ($n=641$), and from DFO Ecosystems and Ocean Science Sector surveys in 2006 \citep{german2006} ($n=2044$), and 2010 \citep{german2010} ($n=3181$) using the Towcam underwater imaging platform. These images included whole-image labels describing the dominant visible substrate type, some of which additionally included detailed comments describing the proportion of cover for multiple substrate types. A separate contribution from the Habitat Ecology Section at BIO comprised \num{1262} images from coastal eelgrass and macroalgae surveys along the Eastern Shore of Nova Scotia between 2019 and 2020 \citep{dfoeelgrass}. These images were extracted from video footage captured by a GoPro HERO7 (1080p or 2.7k resolution) deployed from a drop-down platform for passive drifts at 269 sites. Substrate labels were provided at the whole-image level according to the Coastal and Marine Ecological Classification Standard (CMECS) \citep{cmecs}, as were labels for particular biota, including macroalgae and seagrasses. Finally, the DFO Deep-sea Ecology Program at the Institute of Ocean Sciences (IOS), British Columbia, contributed data collected during the 2018 Northeast Pacific Seamount Expedition using the ROV Hercules. Northeast Pacific Seamount Expedition Partners and Ocean Exploration Trust collected imagery at SGaan Kinghlas-Bowie, Explorer, and Dellwood Seamounts off the west coast of Canada in 2018. Video frames were extracted every 10 seconds for analysis, and \num{16247} were included here. Labels were provided for some images describing the primary substrate type and also the ``biotope'' observed, which broadly describes the benthic community and/or habitat context (e.g. coral garden, vertical wall, sponge ground). Some images overlapped and were thus not originally labelled; in such cases, neighbouring image labels were interpolated where not initially assigned due to overlap with other images. 

\subsubsection{NRCan}

Natural Resources Canada (\href{https://www.nrcan.gc.ca/}{NRCan}) is a federal organization responsible for managing and researching a range of natural resources at the national scale. NRCan makes data freely available via the \href{https://open.canada.ca/en}{Canada Open Government Portal}. The NRCan/GSC Seabed Photo Collection was acquired for this project, which includes \num{20260} images recorded from \num{1804} camera stations across 78 expeditions distributed throughout the waters surrounding Canada. These photographs were collected between 1965 and 2015 using a range of equipment; photographs taken before 1978 were in greyscale, and after 1978 in colour. Photographs before 2000 were collected using film and after 2004 were digital, with both used in the interim. \num{3767} of the photographs were annotated with verbose descriptions of either geological features, biological contents, or both. These descriptions were parsed in order to apply whole-image substrate and biota labels (see \textit{Data management}, \autoref{s:Data-management}, below).
The full \href{https://ftp.maps.canada.ca/pub/nrcan_rncan/Seas_Mer/SeabedPhotoCollection_CollectionPhotosFondsMarins/GSC_Seabed_Photo_Collection.gdb.zip}{list of expeditions} associated with this dataset was obtained along with URLs of corresponding metadata CSV files in GeoDataBase format from the NRCan FTP server. The GeoDataBase file was processed with \href{https://pypi.org/project/geopandas/}{geopandas}, and CSV files were downloaded for each expedition location (URLs were manually corrected for expedition \verb|82FOGO-ISLE|, for which the CSV files were available at URLs containing the string \verb|82FOGO_ISLE| instead).
These CSV files, containing URLs for individual images from the expeditions, were merged together. The year of acquisition was inferred from the expedition name, and columns were renamed to match the standardized dataset format. Sample images were inspected from each expedition to verify their appropriateness. All images from expedition 71014 consisted of collages formed of 2--6 individual photographs, and were excluded.

\subsubsection{NGU}

The Geological Survey of Norway (\href{https://www.ngu.no/en}{NGU}) is responsible for national geological mapping and research, including marine applications. NGU contributed \num{50290} images to this project, which were extracted from 581 underwater video transects acquired during six cruises. These were carried out between 2010 and 2017 in coastal areas and fjords of Norway (Astafjorden, Frohavet, Søre Sunmøre, Sogn og Fjordane, Ofoten, Tysfjorden, and Tjeldsundet), as part of several ``Marine Base Maps'' projects. The videos were acquired using a camera rig towed near the seafloor (\SIrange{0}{200}{\metre} depth) from the NGU research vessel \textit{Seisma}. The 2010 cruises (codes 1002 and 1007) used a 720x480 digital video camera, while all the other cruises (codes 1408, 1508, 1511, and 1706) used a higher-resolution GoPro HERO3+. The images were obtained by extracting one video frame every 10 seconds of video footage.

\subsubsection{MGDS}

The Marine Geoscience Data System (\href{https://www.marine-geo.org/}{MGDS}) is a data repository that offers public access to a curated collection of marine geophysical data products and complementary data related to understanding the formation and evolution of the seafloor and sub-seafloor. MGDS provides tools and services for the discovery and download of data collected throughout the global oceans produced primarily by researchers funded by the U.S. National Science Foundation. Six datasets were obtained from MGDS, in collaboration with the Lamont-Doherty Earth Observatory at Columbia University. Four of these were collected from the Long Island Sound Estuary in 2012 and 2013 using the United States Geological Survey (\href{https://www.usgs.gov/}{USGS}) Seabed Observation and Sampling System (SEABOSS), Integrated Seafloor Imagery System camera sled, and the Kraken2 ROV \citep{zajac2020}. One dataset was obtained from the East Pacific Rise Spreading Center during the 2011 Atlantis expedition, using an Insite Scorpio Digital Camera mounted on the ROV Jason II. The final dataset was acquired by the Schmidt Ocean Institute (SOI) during the 2020 R/V \textit{Falkor} expedition FK200429 off the northeast coast of Australia. Here, the ROV SuBastian was mobilized and images were obtained using a Subsea Systems and Inc. Z70 Digital Camera. All datasets from MGDS were manually reviewed and filtered to remove surface images (e.g. on the research vessel) and duplicates.

\subsubsection{NOAA}

The U.S. National Oceanic and Atmospheric Administration (\href{https://www.noaa.gov/}{NOAA}) is a federal science institution that conducts extensive marine research. NOAA hosts diverse collections of environmental data that are made available to the public. Benthic images were sourced from the NOAA data repository for addition to the BenthicNet dataset. Candidate data were identified using the NOAA \href{https://data.noaa.gov/onestop/}{OneStop portal}, using the search strings ``benthic'', ``habitat'', ``image'', ``camera'', and ``photograph''. Datasets returned not containing image files were rejected. The remainder were reviewed manually, and datasets were additionally rejected that did not meet quality or content standards. Reasons for rejection included substantial proportions of non-benthic images (e.g. above-water, pelagic, individual animals, air photos), partial or full scene obstruction by non-benthic objects (e.g. equipment, ROV/AUV parts), highly inconsistent image content or quality, and incoherent dataset or metadata formatting (e.g. unorganized collections of various types of data, metadata not readable via script). Datasets were also excluded that did not meet the metadata requirements of this project --- namely, those lacking metadata entirely, or lacking geographic locations for images. Where the latter occurred, efforts were made to estimate image locations using available information; for example, by assigning general study site coordinates to images, or by assigning the mean geographic centre of other images at the study site. Datasets that were otherwise suitable for inclusion were generally not rejected due to poor image quality or low resolution alone. All datasets were subjected to the quality control checks listed previously before downloading for inclusion in the BenthicNet collection, and columns were renamed to match the standardized dataset format. Several datasets included labels associated with the National Coral Reef Monitoring Program (NCRMP) describing the benthic cover, which primarily comprised coral taxa and substrate labels applied to both whole-images and points. These labels were retained.

Additional data was contributed by the NOAA Northeast Fisheries Science Center (\href{https://www.fisheries.noaa.gov/about/northeast-fisheries-science-center}{NEFSC}). These included benthic images from Georges Bank, the Mid-Atlantic Bight, and off the coast of Cape Cod ($n=2240$). Image surveys were conducted in 2015 using the NOAA HabCam benthic imaging platform. Whole-image labels were provided indicating the primary and secondary substrate types, and also the presence of certain taxa (mussels, \textit{Didemnum} tunicates, bryozoans).

\subsubsection{USGS}

The United States Geological Survey (USGS) is a federal organization that conducts earth science research and provides public geoscience information and data. A series of unlabelled benthic image datasets were retrieved from the USGS \href{https://data.usgs.gov/datacatalog}{Science Data Catalogue}. Several of these were initially discovered from review of the scientific literature \citep{lidz2013, zawada2015}, and the remainder were discovered by querying the repository using the search strings ``benthic'', ``habitat'', ``image'', ``camera'', and ``photograph''. Candidate datasets were screened using the same methodology as outlined above for data retrieved from the NOAA repository. Datasets were rejected that did not contain images, contained non-benthic images, were largely obstructed by non-benthic objects, or were formatted incoherently. Where precise image locations were not provided, estimates were obtained using the mean centre of the study site bounding box coordinates. All candidate datasets were subjected to the quality control checks listed previously and columns were renamed to match the standardized dataset format.

\subsubsection{USAP-DC}

The U.S. Antarctic Program Data Center (\href{https://www.usap-dc.org/}{USAP-DC}) is funded by the U.S. National Science Foundation and is a domain repository for U.S. Antarctic Research data from all disciplines. Five unlabelled datasets were obtained from USAP-DC. These were discovered from the USAP-DC website using the search strings ``benthic'', ``habitat'', ``image'', ``camera'', and ``photograph''. Datasets were screened using the methodology described for the NOAA and USGS repositories. Additionally, some images that did not depict the seabed (e.g. pictures on the boat deck) were manually omitted. The mean centre of the study site bounding boxes were used to estimate image locations where precise positioning was not provided. These were checked for quality using the methodology described previously and columns were renamed to match the standardized dataset format.

\subsubsection{AADC}

The Australian Antarctic Data Centre (\href{https://data.aad.gov.au/}{AADC}) is a long-term repository for Australia's Antarctic data. This data is freely and openly available for scientific use. Two datasets were obtained for this project from the AADC data portal. Seafloor images ($n=203$) from the Sabrina slope, East Antarctica, were collected in 2017 over four transects during survey ``IN2017\_V01'' using the Australian CSIRO Marine National Facility's Deep Tow Camera \citep{sabrina2020}, and were downloaded along with associated metadata from AADC. These included whole-image labels indicating the substrate type coverage and the presence of biota; the former were retained here. Additionally, Geoscience Australia and the Australian Antarctic Division collected underwater photographs in 2011 at 97 sites in the Mertz Glacier region of Antarctica \citep{vms2017}, and \num{1853} images were acquired for this project. Images and metadata from both datasets were checked for quality and formatted for standardization with the BenthicNet compilation.

\subsubsection{SQUIDLE+}
 
\href{https://squidle.org/}{SQUIDLE+} is an online tool for managing, exploring, and annotating images and video of the seafloor. It also serves as a global repository, containing standardized records for images collected by different groups around the world. SQUIDLE+ is a living product that is updated continuously with new images and labels. A snapshot of the images available on SQUIDLE+ was acquired on April 13, 2023. The SQUIDLE+ \href{https://squidle.org/api/help?template=api_help_page.html}{web API} was used to download the records for every image on SQUIDLE+, totalling \num{9166472} at that time. The paginated download was joined together and merged into a single CSV file, and columns were renamed to match our standardized format for the compilation.

Several of the large individual SQUIDLE+ datasets in this collection additionally included publicly accessible image annotations. These included Australia’s Integrated Marine Observing System (\href{https://imos.org.au/}{IMOS}), which distributes oceanographic data from a consortium of Australian institutions that is freely and openly available to the scientific community. This data included a large number of images collected by the IMOS AUV Facility, notably, using Sirius and Nimbus AUVs. IMOS images available from SQUIDLE+ were cross-referenced with data entries from the Australian Ocean Data Network (\href{https://portal.aodn.org.au/}{AODN}) portal for this project. Labelled images were also provided by the Reef Life Survey (\href{https://reeflifesurvey.com/}{RLS}) \citep{RLS, edgar_2020}, which is a global citizen science program that trains SCUBA divers to conduct underwater visual surveys of shallow reef biodiversity in temperate and tropical reef habitats, typically between \qtyrange[range-phrase=~--~]{2}{20}{\metre} depth. Divers capture approximately 20 images per survey using an underwater camera positioned approximately \SI{50}{\cm} from the substrate, and images vary in resolution and quality due to camera configuration and environmental conditions. The Schmidt Ocean Institute (\href{https://schmidtocean.org/}{SOI}) is a non-profit foundation established to advance global oceanographic research that hosts a large labelled image collection on SQUIDLE+. Deployed from the SOI R/V \textit{Falkor}, the ROV \textit{SuBastian} has collected high resolution images from waters around the world, including the deep ocean. All oceanographic data collected by the SOI are made openly available for research purposes. The National Environmental Science Program (\href{https://www.nespmarine.edu.au/}{NESP}) Marine Biodiversity Hub \citep{nesp} has also provided a large labelled image dataset. This project aims to provide foundational science for conservation in Australian and provides data openly in support of marine research. Each of the above datasets included sub-image point labels identifying underlying physical or biological elements according to the CATAMI scheme \citep{catami}. Finally, the image dataset presented by \citet{yamada2021} collected via AUV from the Southern Hydrate Ridge was downloaded from a separate SQUIDLE domain, \href{https://soi.squidle.org/}{SOI SQUIDLE+}, along with point annotations describing substrate or biotic elements according to a site-specific scheme.

\subsubsection{FathomNet}

\href{http://fathomnet.org}{FathomNet} \citep{katija_fathomnet} is an open-source underwater image database with global scope operated by the Monterey Bay Aquarium Research Institute (\href{https://www.mbari.org/}{MBARI}). FathomNet is soliciting contributions from around the world to develop a large open-source database of images that may be used to develop artificial intelligence algorithms, with a focus on identifying marine species. Like SQUIDLE+, FathomNet is a living product that is updated continuously. We used the \href{https://pypi.org/project/fathomnet/}{FathomNet Python API} to download a snapshot of the images available on FathomNet as of April 6, 2023. The code for this API call is provided in \aref{c:fathomnet_code}. At the time of downloading, these images were primarily acquired from Pacific Waters around California, Western USA. Records were partitioned into ``sites'' based on the directory structure in the URL. Where not available in the record itself, timestamps were extracted from image names, where possible. Columns were renamed to match our standardized format. Many of the images were annotated with bounding boxes around animals and other concepts appearing in the images. However, annotations were available only under a No Derivatives license (\href{https://creativecommons.org/licenses/by-nd/4.0/}{CC~BY-ND~4.0}), which prohibited conversion to other schemes and formats. All FathomNet annotations were thus discarded.

\subsubsection{PANGAEA}

\href{https://www.pangaea.de/}{PANGAEA} is an open access repository aimed at archiving, publishing and distributing georeferenced data from earth system research, hosting 678 projects and \num{408811} datasets from various fields at the time of writing. We searched and retrieved benthic image datasets from PANGAEA with a combination of API calls and web-scraping, then pruned the resulting datasets and reformatted them. The \href{https://pypi.org/project/pangaeapy/}{pangaeapy} Python package \citep{pangaeapy} was used to interface with the PANGAEA library. Using the \verb|PanQuery| API,  PANGAEA was searched for 20 queries with various combinations of benthic environment related keywords to find photographs of the seafloor (see \aref{a:pangaea-search} for complete list). The \verb|PanDataSet| API was used to retrieve the metadata for the dataset IDs identified in these searches. Some IDs corresponded to dataset series, which list multiple child datasets. In these cases, all child datasets were retrieved. Some datasets were available in tabular format, and were downloaded directly. Other datasets were paginated, with images hosted on webpages on PANGAEA; these could not be downloaded with the API and were scraped with a custom webscraper using the \verb|BeautifulSoup4| and \verb|request| libraries.

All datasets returned by this search as of January 1, 2024 were downloaded and results were filtered as follows. 
(1) Datasets that did not possess a column containing the word ``url'' or ``image'' that was populated by hyperlinks to files in an image format (TIFF, JPEG, PNG, BMP, CR2) were removed to enable automation of the data acquisition process. It was not possible to verify any ZIP file would contain images without downloading it, and was impractical to automatically associate metadata with the images within a ZIP file of unknown structure. Datasets with images only available to download as a ZIP file were thus discarded.
(2) False positives from the search (datasets comprising imagery not of the seafloor) were filtered out by removing datasets with titles containing undesired keywords appearing in a manually curated blacklist (e.g. ``aquarium'', ``meteorological observations'', ``sea ice conditions'', ``do not use'').
(3) URLs for images consisting of maps, other dataset summary figures, and inappropriate photo subjects were filtered out by removing data hosted on PANGAEA subdomains dedicated to subjects such as maps, projects, publications, sea ice, and satellite imagery.
(4) Images were removed where the URL contained text indicating the subject matter was otherwise inappropriate (e.g. ``dredgephotos'', ``grabsample'', ``core'', ``aquarium'', ``divemap'').
Finally, the columns in the CSV files were renamed to our standardized format. Details for individual datasets are provided with the BenthicNet metadata \citep{BenthicNet_FRDR}.

Several of the datasets obtained from PANGAEA included thematic labels corresponding to benthic images. Many of these were labels of specific biota identified to the highest possible taxonomic resolution, some of which included estimates of percentage cover of each organism in the image. Several of the latter datasets comprised experimental growth plates harbouring the labelled biota. Some datasets additionally included labels for trash and anthropogenic debris. All labels were dropped where datasets indicated usage of machine-assisted annotation instead of manual annotation. Finally, additional point labels were obtained for datasets from the Great Barrier Reef Marine Park, eastern Australia, collected for habitat mapping purposes by the University of Queensland Remote Sensing Research Centre. These datasets comprised quadrat images collected via snorkel and diving from over 100 reefs throughout the Great Barrier Reef Marine Park \citep{roelfsema2018_pangaea, Roelfsema_2020, roelfsema2018_heron}. Points were labelled according to a custom scheme used for these projects at the Great Barrier Reef that describe biotic and abiotic elements found within the reef. Additional labels were also provided indicating the biotic functional group, and a simplified classification scheme applicable to a global context.

\subsubsection{XL Catlin Seaview Survey}

The \href{https://www.catlinseaviewsurvey.com/}{XL Catlin Seaview Survey} was a large-scale project undertaken between 2012--2018 to document and study the status of coral reefs globally using underwater imagery. Surveys focused on shallow reefs typically around \SI{10}{\metre} depth and comprised linear transects ranging between \SIrange{1.6}{2}{\kilo\metre} in length. Downward-facing seabed images of approximately \SI{1}{\metre\squared} were acquired using Canon 5D MII cameras mounted on a self-propelled diver-operated platform called the ``SVII'' \citep{gonzalez2014catlin, gonzalez_2016}. Data from the project is made openly available for further scientific research. For this project, \num{1082452} images from 860 surveys organized into 22 regional datasets were downloaded from the University of Queensland \href{https://espace.library.uq.edu.au/view/UQ:734799}{data repository}. Tabular data providing image metadata was also acquired in CSV format, including image point labels identifying biotic and abiotic elements using the global scheme applied above for the Great Barrier Reef mapping projects. The metadata were renamed and formatted to match the standardized BenthicNet compilation.

\begin{table}[tbh]
  \centering
  \caption{%
\textbf{Summary of BenthicNet data sources} including the number of images in BenthicNet-11M (Full collection), BenthicNet-1M (Subsampled), and BethicNet-L (Labelled). Further details on the individual datasets are provided within the BenthicNet metadata \citep{BenthicNet_FRDR}
}
\label{tab:datasets}
\centerline{
\small
\begin{tabular}{llrrrrr}
\toprule
               &           &                         &                      & \multicolumn{3}{c}{\textnumero{} Samples} \\
\cmidrule(lr){5-7}
Source     & Region    & \textnumero{} Datasets  & \textnumero{} Sites  & Full collection & Subsampled & Labelled \\
\midrule
\textit{Online Repository/Collection} \\
AADC                & Antarctic & \num{   2} & \num{   86} & \num{   2056} & \num{  2024} & \num{  203} \\
Catlin Seaview      & Global    & \num{  22} & \num{  861} & \num{1082452} & \num{283674} & \num{11346} \\
FathomNet           & W. USA    & \num{   8} & \num{ 3381} & \num{  68908} & \num{ 58196} & \num{    0} \\
MGDS                & Global    & \num{   6} & \num{   32} & \num{  15023} & \num{  6154} & \num{    0} \\
NOAA (via OneStop)  & USA       & \num{  18} & \num{  526} & \num{  73019} & \num{ 40714} & \num{ 4543} \\
NRCan               & Canada    & \num{  78} & \num{ 1804} & \num{  23855} & \num{ 18851} & \num{ 3595} \\
PANGAEA             & Global    & \num{1191} & \num{ 1196} & \num{ 764924} & \num{236968} & \num{40204} \\
SQUIDLE+            & Global    & \num{ 691} & \num{14187} & \num{9166472} & \num{608576} & \num{85387} \\
USAP-DC             & Antarctic & \num{   5} & \num{   27} & \num{   4144} & \num{  2886} & \num{    0} \\
USGS                & USA       & \num{   5} & \num{   38} & \num{ 104155} & \num{  7035} & \num{    0} \\
\midrule
\textit{Individual Contributions} \\
4D Oceans           & E. Canada & \num{   2} & \num{  274} & \num{   3008} & \num{  2715} & \num{ 3000} \\
DFO (BIO)           & E. Canada & \num{   6} & \num{  381} & \num{   7773} & \num{  5981} & \num{ 7762} \\
DFO (IOS)           & W. Canada & \num{   7} & \num{    9} & \num{  16247} & \num{  1993} & \num{10106} \\
EAC                 & E. Canada & \num{   1} & \num{    7} & \num{   1220} & \num{  1015} & \num{  886} \\
Hakai Institute     & W. Canada & \num{   2} & \num{   45} & \num{   4735} & \num{  3609} & \num{ 1697} \\
HAL                 & Jamaica   & \num{   1} & \num{    1} & \num{    505} & \num{   505} & \num{    0} \\
LaboGeo/UFES
                    & E. Brazil & \num{   1} & \num{  359} & \num{    359} & \num{   287} & \num{  359} \\
MUN                 & Arctic    & \num{   4} & \num{  135} & \num{  10691} & \num{  6403} & \num{10687} \\
NGU                 & Norway    & \num{   4} & \num{  580} & \num{  50290} & \num{ 50275} & \num{    0} \\
NOAA (NEFSC)        & N.E. USA  & \num{   1} & \num{    2} & \num{   2240} & \num{  2065} & \num{ 2240} \\
SEAM                & E. Canada & \num{   3} & \num{  284} & \num{   6811} & \num{  5170} & \num{ 6673} \\
\midrule
Total               & Global    & \num{2058} & \num{24215} &\num{11408887} &\num{1345096} & \num{188688} \\
\bottomrule
\end{tabular}
}
\end{table}

\subsection{Data management}
\label{s:Data-management}

In total, \num{11408887} images were collected from the sources described above (see \textit{Data compilation and quality control}, \autoref{s:Data-compilation-and-quality-control}). The greatest discrepancy within this collection was the presence of image labels. Of all the images acquired, only \num{188688} included labels corresponding to visible benthic elements. The presence and composition of labels in many ways determines the utility of the dataset; labels enable training and validation for supervised modelling tasks, such as localized species or substrate identifications \citep{jacket2023, piechaud2019}, or bottom type classification \citep{diegues2018}. There are several ways, though, that unlabelled data may still be utilized using unsupervised \citep{yamada2021}, semi-supervised \citep{arosio2023,Ouali2020,vanEngelen2020,Yang2023}, and self-supervised \citep{huang2022, SSLCookbook, SimCLR, MoCo-v1, BarlowTwins, BYOL, masked_autoencoder} approaches. To facilitate a range of potential applications, we consider the dataset in two ways hereafter: the full set of images without their labels (unlabelled; BenthicNet-11M) and the set of labelled images (BenthicNet-Labelled).

\subsubsection{Labelled data}
\label{s:Labelled-data}

In order to increase the utility of the compiled data, and to facilitate validation of models trained on it, image labels from all datasets were translated to the CATAMI classification scheme \citep{catami} (version 1.4), which spans both substrate and biota categories. Biota labels were additionally mapped to the World Registry of Marine Species (WoRMS) taxonomy \citep{WoRMS}.

\begin{table}[tbh]
\caption{%
\textbf{Examples of original image labels translated to hierarchical labels} according to CATAMI v1.4 and WoRMS. Some original labels indicated both substrate and biota, while others indicated only one of these. For biota, some original labels provided more morphological detail and others more taxonomic; as much detail was retained as possible in both the CATAMI Biota and WoRMS taxonomic translations, respectively.
}
\label{tab:catami_ex}
\centering
\small
\begin{tabular}{lllcl}
\toprule
& \multicolumn{2}{c}{CATAMI} & \multicolumn{2}{c}{WoRMS}\\
\cmidrule(l){2-3} \cmidrule(l){4-5}
\multicolumn{1}{c}{Original} & \multicolumn{1}{c}{Substrate} & \multicolumn{1}{c}{Biota} & \multicolumn{1}{c}{AphiaID} & \multicolumn{1}{c}{Taxonomy} \\
\midrule
Cobble & \begin{tabular}[c]{@{}l@{}}Substrate \\\hspace{1mm}↳ Consolidated (hard) \\\hspace{4mm}↳ Cobbles\end{tabular} & \noval{} & \noval{} & \noval{} \\
\midrule
Mud and tube worms & \begin{tabular}[c]{@{}l@{}}Substrate\\ \hspace{1mm}↳ Unconsolidated (soft)\\ \hspace{4mm}↳ Sand / mud (\textless{}1mm)\\ \hspace{7mm}↳ Mud / silt (\textless{}64um)\end{tabular} & \begin{tabular}[c]{@{}l@{}}Worms\\ \hspace{1mm}↳ Polychaetes\\ \hspace{4mm}↳ Tube worms\end{tabular} & \href{https://www.marinespecies.org/aphia.php?p=taxdetails&id=883}{883} & \begin{tabular}[c]{@{}l@{}}Annelida\\ \hspace{1mm}↳ Polychaeta\end{tabular}\\
\midrule
\begin{tabular}[c]{@{}l@{}}Hard Coral:\\Non hermatypic:\\Free living (Fungia etc)\end{tabular} & \noval{} & \begin{tabular}[c]{@{}l@{}}Cnidaria\\ \hspace{1mm}↳ Corals\\ \hspace{4mm}↳ Stony corals\\\hspace{7mm}↳ Solitary\\\hspace{10mm}↳ Free living\end{tabular} & \href{https://www.marinespecies.org/aphia.php?p=taxdetails&id=1363}{1363} & \begin{tabular}[c]{@{}l@{}}Cnidaria\\ \hspace{1mm}↳ Hexacorallia\\ \hspace{4mm}↳ Scleractinia\end{tabular} \\
\midrule
\genus{Pocillopora} sp. & \noval{} & \begin{tabular}[c]{@{}l@{}}Cnidaria\\ \hspace{1mm}↳ Corals\\ \hspace{4mm}↳ Stony corals\end{tabular} & \href{https://www.marinespecies.org/aphia.php?p=taxdetails&id=206938}{206938} & \begin{tabular}[c]{@{}l@{}}Cnidaria\\ \hspace{1mm}↳ Hexacorallia\\ \hspace{4mm}↳ Scleractinia\\ \hspace{7mm}↳ Pocilloporidae\\ \hspace{10mm}↳ \genus{Pocillopora}\end{tabular} \\
\bottomrule
\end{tabular}
\end{table}
 
 Images were originally labelled according to a range of different established and bespoke schemes, yet a large number of these (for example, the labelled data acquired from SQUIDLE+) were readily available as CATAMI labels. Additionally, the qualities of CATAMI provide a flexible framework that may accommodate translation and integration of a broad range of other labelled data. First, CATAMI supports labels for multiple classes of benthic features, including ``branches'' for both biota and physical elements such as substrate, bedforms, and relief. This enables the translation of a range of labelled datasets that were initially collected for a variety of different purposes. Second, the labels within these branches are hierarchical. This means that objects may be labelled at different or even multiple levels of detail depending on the quality of the data, the confidence of the analyst, or the requirements of a particular application. This characteristic is critical for the translation of the multi-source data compiled here, which were initially analyzed at a range of thematic (e.g. taxonomic) resolutions for different purposes. Finally, CATAMI implements labels that are designed to be visually recognizable from image data. At a coarse level, these may distinguish broad groups or phyla of biota, but at finer levels, where identifying individual genera or species may become difficult using image data alone, morphological labels may be applied. These describe the size, shape, colour, and growth form of an organism, which may be recognizable where the taxonomy is ambiguous. Detailed taxonomic labels often require specialized knowledge or biological expertise, whereas morphological labels enable collaborative classification and translation of imagery by non-experts. 

Translation of all image labels to the CATAMI scheme was performed by a team of BenthicNet collaborators (e.g. \autoref{tab:catami_ex}). All unique labels were extracted for each labelled image dataset in turn, which included cases of a single label indicating one benthic feature (e.g. sediment type or biota), a single label indicating multiple features (e.g. sediment type and biota), or multiple labels for different features (e.g. one for sediment type, one for biota). Each unique label for each dataset was translated to its closest CATAMI equivalent(s), maintaining the hierarchical level of the original data as closely as possible. In some cases, additional information within the metadata such as comments or auxiliary labels were used to complete the translation. Some annotations were provided in schemes that extend versions of CATAMI, such as the Australian Morphospecies Catalogue, which provides more precise morphological detail for the shape of sponges, for example. Where this was done systematically and with more than 10 samples, we extended our scheme by adding child nodes to correspond to the increased level of morphological detail. Some annotations included man-made objects, such as trash or cables, which fell outside the scope of the CATAMI scheme, but which may have value toward monitoring the anthropogenic impact on benthic habitats. Thus we also added an additional Anthropogenic branch to the hierarchy to cater to these annotations. We include fields for CATAMI modifiers that indicate additional information about image labels, such as whether organisms are bleached or dead, or their colours, where available.

Some datasets provided taxonomic labels of biota at a high level of detail (genus or species level). To retain this information, taxonomic biota labels were additionally assigned an AphiaID from the World Registry of Marine Species (WoRMS). Where detailed taxonomic labels could not be determined, remaining biota annotations (e.g. morphological descriptions from CATAMI) were also mapped onto the WoRMS taxonomy at the highest level of specificity possible (typically phylum, class, or order).

In total, there were \num{188688} labelled images, \num{3091158} individual CATAMI labels, and \num{1131391} WoRMS taxonomic labels for the BenthicNet compilation. The counts for each individual label are provided with the dataset hosted on the Canadian Federated Research Data Repository (FRDR) \citep{BenthicNet_FRDR}.

To enable consistent validation and benchmarking between models using the BenthicNet dataset compilation, we propose train and test partitions of the labelled data. Test data were selected according to a partially spatial and stratified procedure in order to ensure representation of a broad range of labels, and to reduce the degree of similarity between test and training partitions caused by spatial autocorrelation. 

The challenge in partitioning the dataset stems from the multi-label nature and imbalanced proportions of labels in BenthicNet. Firstly, the imbalance necessitates careful assignment of rarer labels in the dataset. Additionally, a single image may have any combination of labels across multiple branches of the CATAMI hierarchy. If an image is assigned to test or training partitions due to a particular label, we must consider how the assignment affects other labels on the same image, some of which will be rarer.

Our partitioning process was as follows. We selected the target number of annotations per label to place in the test set as the smaller of 15\% of the number of samples for the most frequently occurring label and 35\% of the samples for the median label. We incrementally added images to train or test partitions one at a time. We considered the available image labels in each iteration, and selected the next label to add to a partition based on the following factors, in order of priority.

\begin{enumerate}
    \item Ensure at least two samples for each label can be placed in the train partition.
    \item Ensure at least 50\% of the samples for each label can be placed in the train partition without using samples within \SI{50}{\metre} of a test sample.
    \item Ensure at least 15\% of the samples for each label can be placed in the test partition without using samples within \SI{50}{\metre} of a train sample.
    \item Ensure no more than 35\% of the samples for each label would be placed in the test partition.
    \item Prioritize the CATAMI label with the fewest remaining images which can be allocated to the train/test partition.
\end{enumerate}

After determining the next label for which a sample will be added to a partition, and the partition to which it should be added, we selected an image bearing that annotation to add to the partition as follows:
\begin{enumerate}
    \item Of the images bearing the desired label, if any images are within \SI{50}{\metre} of an image already in the target partition, randomly select an image from the closest 10\% of those images.
    \item Otherwise, randomly select an image bearing the desired label that is not within \SI{50}{\metre} of an image already allocated to the other partition. Images violating the \SI{50}{\metre} exclusion zone were used if needed to satisfy the minimum populations for each partition described above (50\% in train, 15\% in test).
\end{enumerate}

In practice, this means that partitions grow spatially outwards from initial seed locations, with new locations being seeded at random when needed in order to represent new label classes. Any remaining samples that have not been assigned through this process were allocated to the test partition if within \SI{50}{\metre} of an image already in the test partition, and to the train partition otherwise.

Effectively, test data was selected to prioritize representation of CATAMI labels, and then to minimize spatial overlap with the training data, to the maximum extent possible. Because the computation of this algorithm scales $\bigO(n^2)$ with data size, it was run in parallel on 37 subsets of the data, each corresponding to a different Ecological Marine Unit \citep{sayer2017} (EMU; see \textit{Unlabelled data exploratory analysis}, \autoref{s:Unlabelled-data-exploratory-analysis}). \num{142767} images (75.66\%) were assigned to the training partition and \num{45921} (24.34\%) to test. The code for obtaining training and test partitions of the labelled data is provided at the BenthicNet code repository (see \textit{Code availability}, \autoref{s:Code-availability}, below).

\subsubsection{Unlabelled data}
\label{s:Unlabelled-data}

All images collated in BenthicNet may be used for unlabelled applications, including those images that have labels, thus for the ``unlabelled'' slice of the data we considered all \num{11408887} images. We refer to the full collection as BenthicNet-11M. These images were not distributed uniformly in space; some datasets were characterized by a low sampling intensity, with only a few images per recording site taken manually by divers, while others were densely sampled --- for example, where images were extracted from AUV video. In order to reduce the redundancy of the densely sampled data (thereby also reducing data volume and imbalance) the unlabelled data was subsampled spatially, as described below. We refer to the subsampled dataset as BenthicNet-1M.

The aim of the subsampling procedure was to obtain a manageable unlabelled data volume without reducing the breadth of benthic environments represented. Many datasets indicated which images were collected at the same recording station, or the same camera deployment/transect. We collectively refer to this location annotation as a ``site''. To maximize spatial and thematic diversity of images, subsampling was performed separately for each unique site in the unlabelled datatset.

In order to subsample the data spatially, we first determined a desirable number of images that should be drawn from a given site based on the data density. The base target number of images sought at each site was set to \num{250}, meaning that the subsampling procedure would not reduce the number of images below this number.
Not all component datasets indicated whether images were collected at the same site, despite containing images from multiple distinct locations that would meet our ``site'' criteria. To address this, we automatically detected the number of ``pseudo-sites'' within an annotated site, or within a dataset originally lacking any site labels. Pseudo-sites were determined as clusters of samples at least \SI{1000}{\metre} from each other. The target number of samples was scaled up by the number of pseudo-sites within a labelled site. Some (pseudo-)sites additionally had gaps between them of several hundred metres, which we refer to as ``subsites''. The target number of samples for a site was increased by \num{50} for each subsite within it separated by at least \SI{100}{\metre}.

After determining the target number of images to draw from each site in the unlabelled dataset, the data was subsampled spatially. Sites with fewer than \num{40} samples per pseudo-site were not subsampled. At sites with more than \num{40} images, images were subsampled with a target separation distance of $\Delta=1.25\,\text{m}$ according to the following procedure.
\begin{enumerate}
    \item Add the first image in the dataset.
    \item Continue through the list of images in the dataset (sorted in collection order; i.e. chronologically) until finding the first image at least $\Delta=1.25\,\text{m}$ from the last image added to the dataset.
    \item Add either this image or the previous image in the list, whichever is closest to being a distance $\Delta=1.25\,\text{m}$ from the last image added to the dataset.
    \item From the list of remaining images to consider, remove all images collected within $\Delta/2=0.625\,\text{m}$ of this image.
    \item Return to Step 2; repeat until reaching the end of the dataset.
    \item Add the last image in the dataset if it was at least $\Delta/2=0.625\,\text{m}$ from all other images.
\end{enumerate}
Sites lacking precise coordinate information for each image could not be subsampled spatially. In these cases, sites were subsampled by keeping every $n$-th image (ordered chronologically) at the site to achieve the desired number.

\begin{figure}[tbh]
    \centerline{
        \includegraphics{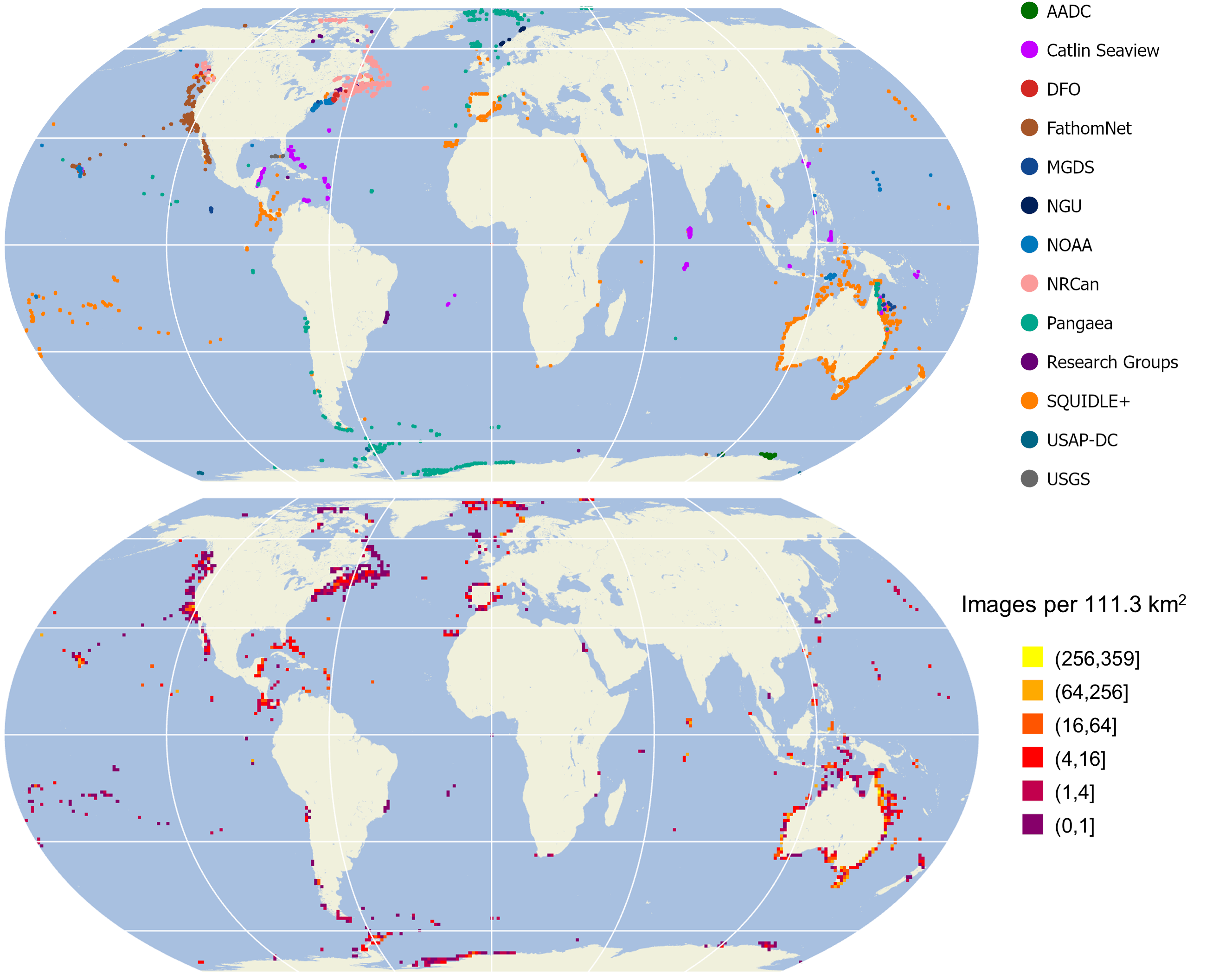}
    }
    \caption{\textbf{Distribution of images after spatial subsampling} projected to Equal Earth. (Top) images according to data source and (bottom) aggregated by their density and scaled logarithmically.}
    \label{fig:scatter}
\end{figure}

Many sites still had more images than their target number of samples after this initial spatial subsampling, so this process was repeated with larger separation distances until the target subsample size was achieved at each site, or a maximum downsampling separation distance of \SI{20}{\metre} was reached. Separation distances were scaled up by factors of 2, 3, 4, 6, 8, 10, 12, 14, or 16 compared to the base subsampling of \SI{1.25}{\metre} target separation to achieve the desired subsample size (i.e. $\Delta=$ \SI{2.5}{\metre}, \SI{3.75}{\metre}, \ldots, \SI{20}{\metre}). The subsampling distance selected (and hence subsampled set of images at that site) was the largest distance that did not reduce the total number of images below the target for the site (250+), determined as described above. The subsampling procedure selected \num{1345096} images (\num{11.8}\% of the total) to be included in the subsampled BenthicNet dataset (\autoref{fig:scatter}), which we refer to as BenthicNet-1M.

\subsubsection{Downloading}

We downloaded images available online by using the Python \href{https://pypi.org/project/requests/}{requests} package.
URLs were retried at least five times if the server was busy.
Images that could not be found at their URL, that were truncated, or which could not be opened after downloading, were removed from the final dataset.
In total, the download process took approximately six months.

\subsubsection{Compression}

Large images were downsampled such that their shortest side was around 512 pixels in length with 10\% tolerance, respecting the original aspect ratio, and then converted to JPEG format for the BenthicNet compilation. The full original uncompressed images are available at URLs provided with the dataset \citep{BenthicNet_FRDR}.

\section{Data Records}

All BenthicNet data, metadata, and models described here are available from the Canadian Federated Research Data Repository (FRDR) \citep{BenthicNet_FRDR}. These include (i) a CSV file with an entry for each image in the subsampled compilation (``BenthicNet-1M''), conforming to the convention presented in \autoref{tab:dataformat}; (ii) a single CSV file with an entry for each label of each image of the labelled compilation (``BenthicNet-Labelled''), conforming to the format presented in \autoref{tab:dataformat_labelled}; (iii) a tarred directory containing each image in BenthicNet-1M and BenthicNet-Labelled (as described in the CSV files above), resized and compressed to JPEG format; (iv) a version of the entire image compilation tarred at the individual ``dataset'' level; and (v) the ResNet-50 model weights resulting from self-supervised training on the entire unlabelled dataset as described in the following \textit{Technical Validation} section, \autoref{s:Technical-Validation}. We additionally include CSV files listing the counts of each individual CATAMI label present in the labelled compilation, and also a list of WoRMS taxonomical labels present within the metadata CSV. The ``image'' directory hosted on FRDR is divided into ``labelled'' and ``unlabelled'' components, which contain the full tarred (\texttt{full\_labelled\_512px.tar}) and individual dataset tars (\texttt{individual\_datasets\_tars}) versions of the images. Both directories are organized by ``dataset'', which contain sub-directories corresponding to the ``site''. Thus, it is possible to query the image compilation using the corresponding CSV metadata fields ``dataset'' > ``site'' > ``image''. A detailed overview and description of the dataset and metadata file structure is provided in the BenthicNet repository \citep{BenthicNet_FRDR} README file (\texttt{00\_Documentation/README.txt}). The metadata and models are available for use without restriction under the Creative Commons Attribution 4.0 License (\href{https://creativecommons.org/licenses/by/4.0/}{CC-BY-4.0}). Most images are available for use without restriction under CC-BY-4.0, except where the original licenses of individual datasets indicate limitations to derivative or commercial uses. The individual licenses for all datasets comprising BenthicNet are retained, which are available along with the metadata within the repository.

\subsection{Data formats}

All unlabelled image metadata were standardized to a common format (\autoref{tab:dataformat}). The datetime field was completed to the highest level of precision possible. Times were converted to UTC where timezones were indicated, and assumed to be UTC otherwise; it is not possible to guarantee all times are in UTC. Missing datetime and coordinate information was imputed everywhere where reasonably possible --- for example, by assigning the geographic mean centre of the image acquisition site where coordinates were missing for some images at a given site. In some cases, any of month, day, hour, minute, and second datetime information was missing, and was imputed to achieve the desired format (\autoref{tab:dataformat}); as a whole, this information should be considered accurate to the year. Labelled images were additionally assigned metadata describing the original and translated CATAMI labels, and WoRMS AphiaIDs (\autoref{tab:dataformat_labelled}). Metadata indicating the pixel location of image labels (i.e. the relative x and y position of the labelled pixel) were retained where provided. 

\begin{table}[tbh]
  \centering
  \caption{%
\textbf{Format for compiled BenthicNet-1M unlabelled image metadata.}
}
\label{tab:dataformat}
\centerline{
\small
\begin{tabular}{llllr}
\toprule
Column          & Contents                              & Data-type & Units             & Coverage \\
\midrule
url             & URL address for this image            & String &                      & 100.00\%\\
source          & Data provider/repository              & String &                      & 100.00\%\\
dataset         & Name of dataset                       & String &                      & 100.00\%\\
site            & Image location name                   & String &                      & 100.00\%\\
image           & Image filename                        & String &                      & 100.00\%\\
latitude        & Latitude (WGS~84)                     & Float  & Decimal degree       &  99.63\%\\
longitude       & Longitude (WGS~84)                    & Float  & Decimal degree       &  99.63\%\\
datetime        & Acquisition date and time (UTC)       & String & YYYY-MM-DD HH:mm:ss  &  99.85\%\\
gebco\_bathymetry& Depth interpolated from GEBCO2022    & Float  & Metres               &  99.63\%\\
emu             & Nearest Ecological Marine Unit        & Integer&                      &  99.63\%\\
\bottomrule
\end{tabular}
}
\end{table}

\begin{table}[tbh]
  \centering
  \caption{%
\textbf{Format for compiled BenthicNet-Labelled image metadata.} Coverage is the fraction of images that have at least one such metadata entry.
}
\label{tab:dataformat_labelled}
\centerline{
\small
\begin{tabular}{llllr}
\toprule
Column            & Contents                              & Data-type & Units            & Coverage \\
\midrule
url               & URL address for this image            & String &                     & 100.00\%\\
source            & Data provider/repository              & String &                     & 100.00\%\\
dataset           & Name of dataset                       & String &                     & 100.00\%\\
site              & Image location name                   & String &                     & 100.00\%\\
image             & Image filename                        & String &                     & 100.00\%\\
latitude          & Latitude (WGS~84)                     & Float  & Decimal degree      & 100.00\%\\
longitude         & Longitude (WGS~84)                    & Float  & Decimal degree      & 100.00\%\\
datetime          & Acquisition date and time (UTC)       & String & YYYY-MM-DD HH:mm:ss & 100.00\%\\
partition         & Train/test split allocation           & String &                     & 100.00\%\\
annotation\_column& Relative x location of labelled pixel & Float  & Fraction of image width   &  53.11\%\\
annotation\_row   & Relative y location of labelled pixel & Float  & Fraction of image height  &  53.11\%\\
original\_label   & Original image label                  & String &                     &  82.10\%\\
catami\_biota     & CATAMI biota label                    & String &                     &  75.59\%\\
catami\_substrate & CATAMI substrate label                & String &                     &  70.30\%\\
catami\_bedforms  & CATAMI bedform label                  & String &                     &   6.87\%\\
catami\_relief    & CATAMI relief label                   & String &                     &   2.46\%\\
catami\_qualifiers& CATAMI label qualifier                & String &                     &  10.82\%\\
colour\_qualifier & Label colour qualifier                & String &                     &   6.52\%\\
bleached          & Whether biota is bleached             & Float  & Values 0 or 1       &  13.44\%\\
dead              & Whether biota is deceased             & Float  & Values 0 or 1       &  25.76\%\\
aphia\_id         & WoRMS taxon AphiaID label             & Integer&                     &  60.73\%\\
gebco\_bathymetry & Depth interpolated from GEBCO2022     & Float  & Metres              & 100.00\%\\
emu               & Nearest Ecological Marine Unit        & Integer&                     &  99.98\%\\
\bottomrule
\end{tabular}
}
\end{table}

\section{Technical Validation}
\label{s:Technical-Validation}

\subsection{Unlabelled data exploratory analysis}
\label{s:Unlabelled-data-exploratory-analysis}

The subsampled BenthicNet dataset contains images from locations around the world (\autoref{fig:scatter}). Several regions are densely sampled --- notably, the entire Australian coast, the Iberian Peninsula, the Norwegian and Greenland Seas, the Eastern Canadian and Northeastern U.S. continental shelves, the Western Canadian and Western U.S. continental shelves, and also some of the Antarctic coast, including parts of the Antarctic Peninsula and Weddell Sea. Other regions are comparatively under-sampled, such as the Indian Ocean including the coast of South Asia, and the South Atlantic including the eastern coast of South America and west coast of Africa. Images collected in the open oceans are more dispersed than those at the continental shelves.

\subsubsection{Environmental heterogeneity}
\begin{figure}[tbh]
    \centerline{
        \includegraphics[scale=1.0]{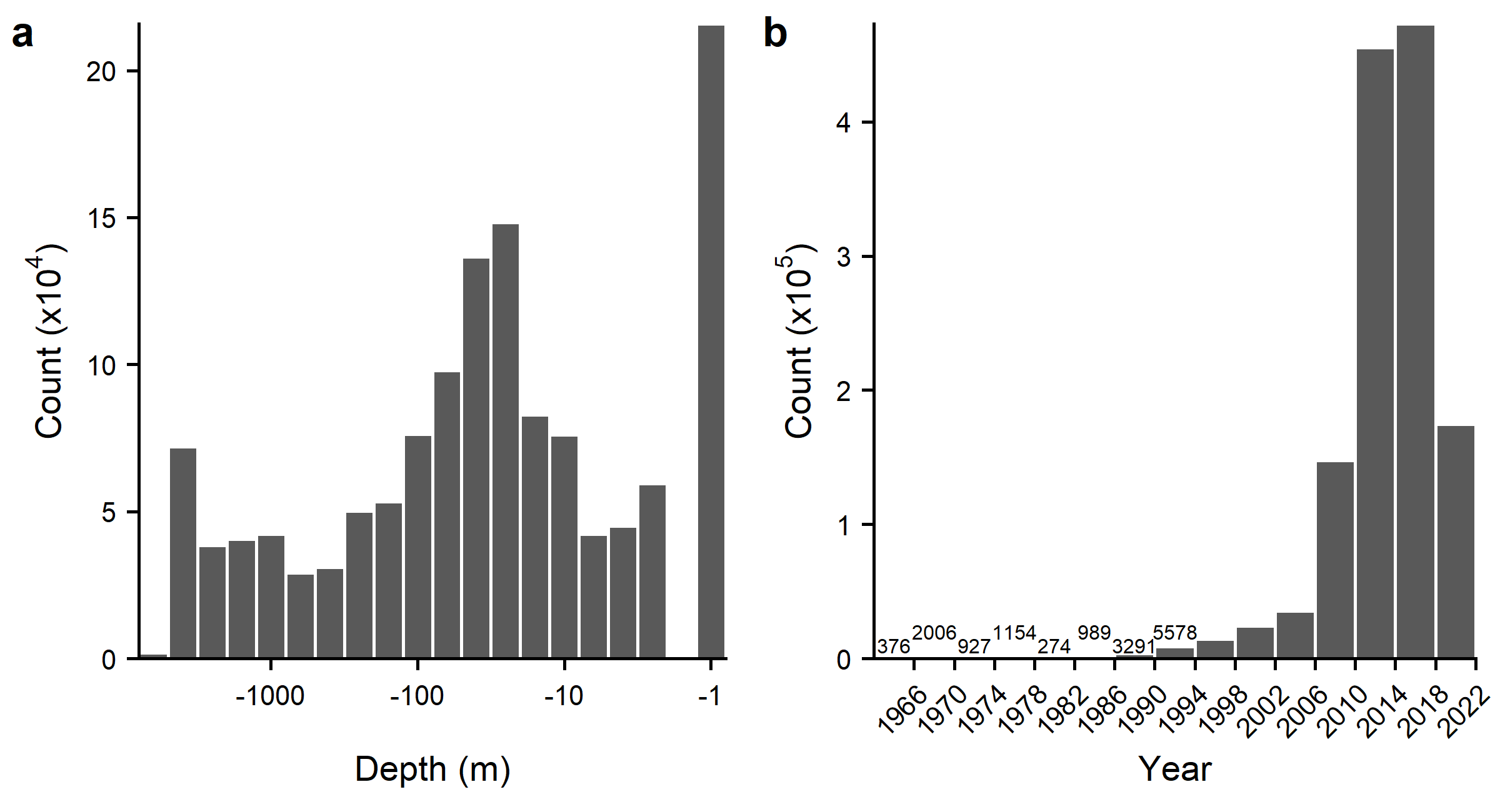}
    }
    \caption{\textbf{Distribution of BenthicNet-1M images} according to (a) (log scale) \textbf{depth} data retrieved from the GEBCO2022 grid \citep{gebco2022} and (b) \textbf{year of acquisition}.}
    \label{fig:depth-distribution}
\end{figure}

\begin{figure}[tbh]
    \centerline{
        \includegraphics[scale=1.0]{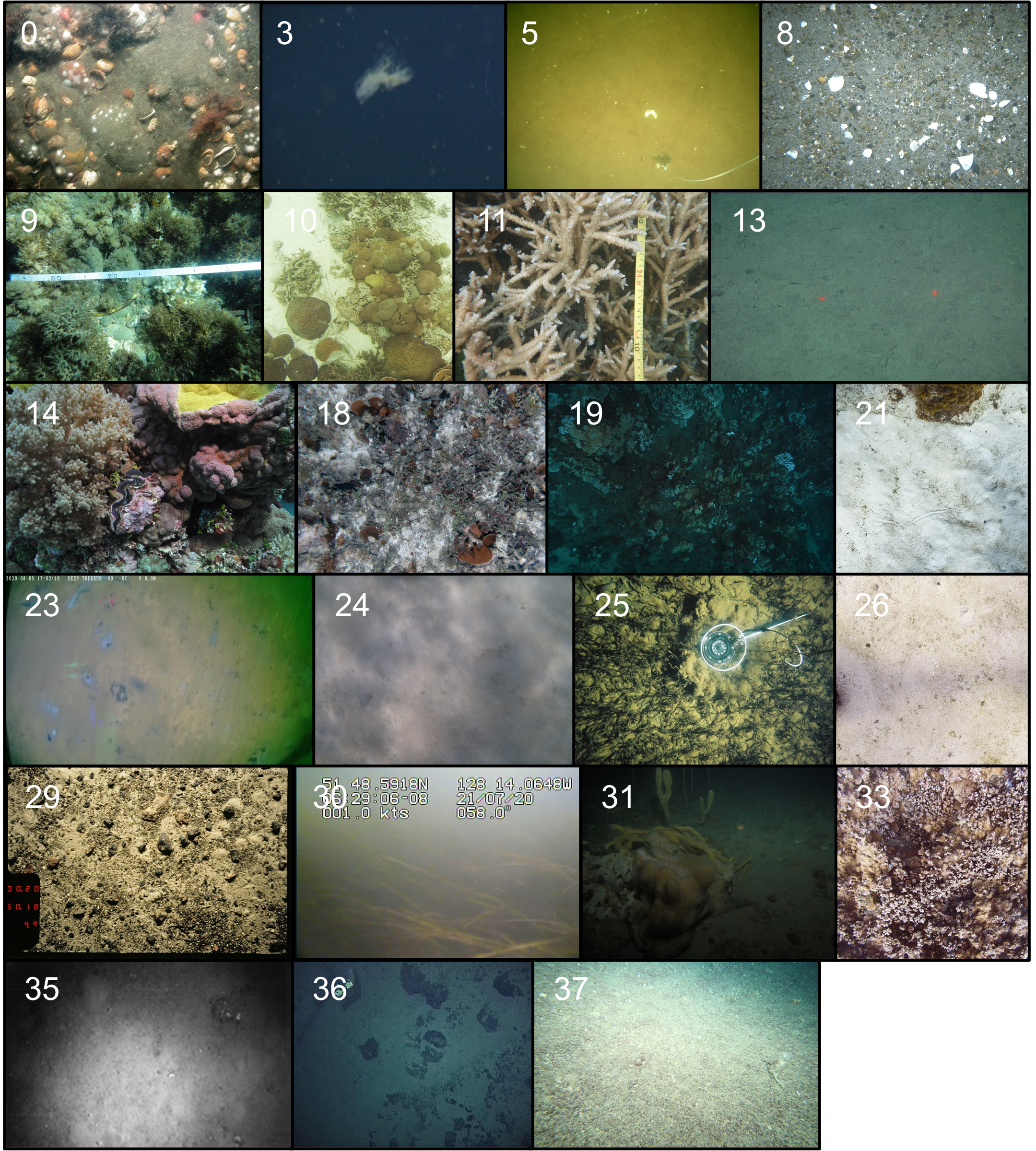}
    }
    \caption{\textbf{Examples of BenthicNet images} from each sampled Ecological Marine Unit (EMU), indicated in the top-left of each image. See \citet{sayer2017} for a full description of the EMU classes.}
    \label{fig:image_emu_eg}
\end{figure}

\begin{figure}[tbh]
    \centerline{
        \includegraphics[scale=1.0]{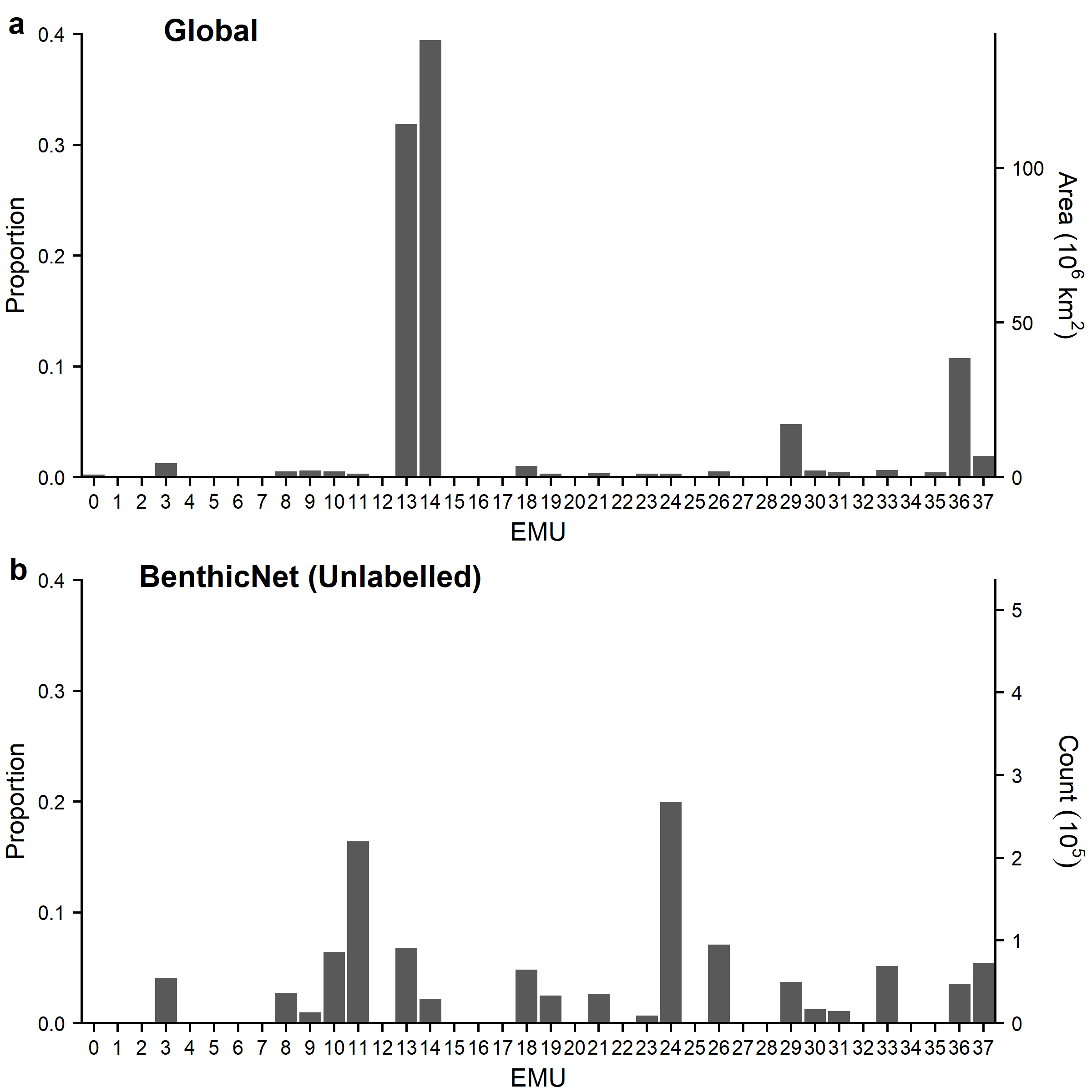}
    }
    \caption{\textbf{Distribution of BenthicNet-1M images according to bottom layer Ecological Marine Units} (EMUs). (a)~Proportion and area of global oceans classified into each EMU. (b)~Proportion of BenthicNet image samples from each EMU. See \citet{sayer2017} for a full description of the EMU classes.}
    \label{fig:emus}
\end{figure}

\begin{figure}[tbh]
    \centerline{
        \includegraphics[scale=1.0]{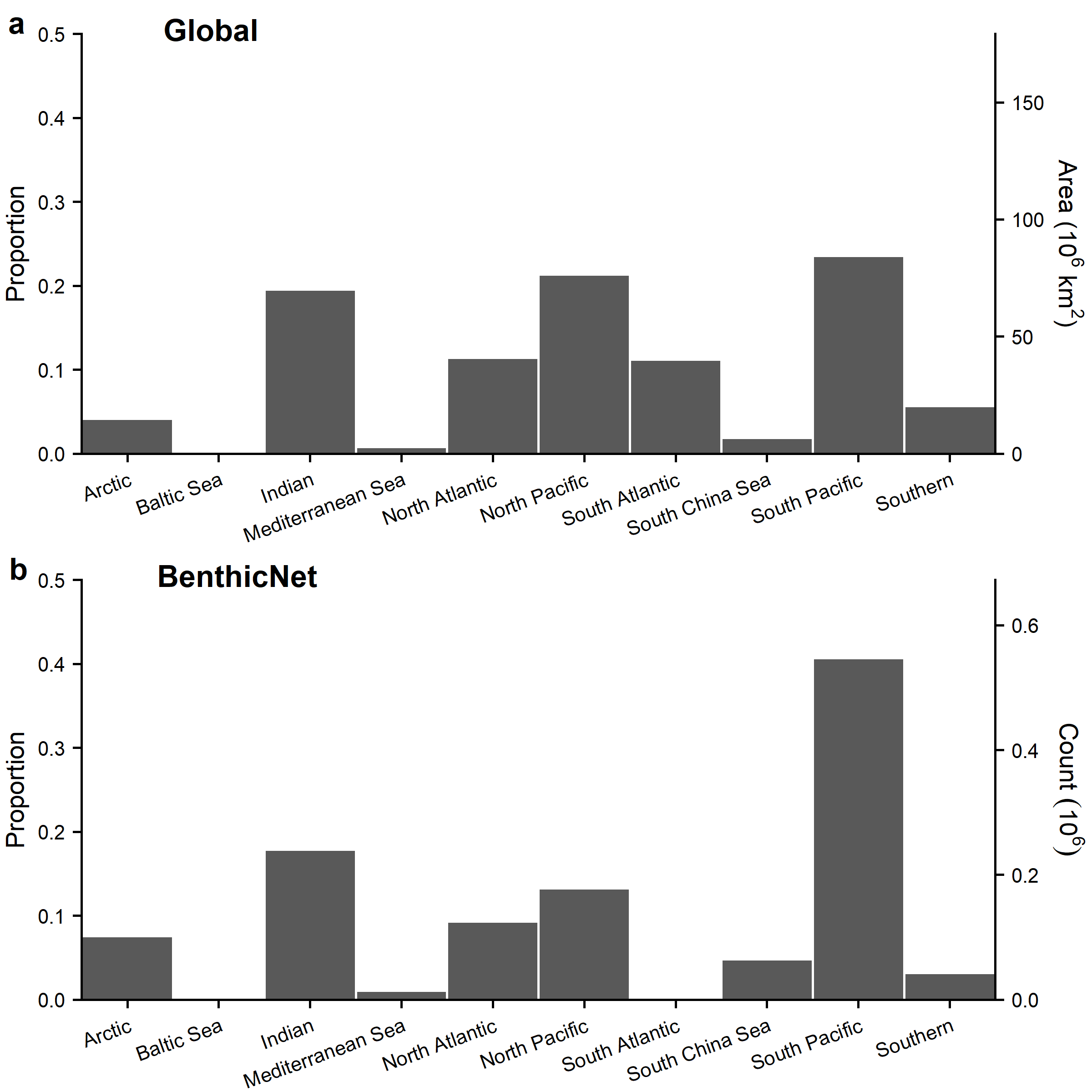}
    }
    \caption{\textbf{Distribution of BenthicNet-1M images according to global ocean basins} \citep{sayer2017}. (a)~Proportion and area of ocean basins. (b)~Proportion of BenthicNet image samples from each ocean basin.}
    \label{fig:oceans}
\end{figure}

Given the spatial heterogeneity in benthic image sampling intensity, it is important and informative to assess the environmental and geographic diversity of these images. Images in the compiled datasets were acquired between 1965--2021 from depths ranging from $<\,\SI{1}{\metre}$ to over \SI{5500}{\metre} (\autoref{fig:depth-distribution}). \Citet{sayer2017} introduced a three-dimensional partitioning of the global oceans into statistical clusters based on a 57-year climatology of physiochemical oceanographic measurements \citep{locarini2013, zwang2013, garcia2014a, garcia2014b}. These 37 ``Ecological Marine Units'' (EMUs) represent a concise and objective summary of global marine environments at \SI{0.25}{\degree} horizontal resolution, and are freely available for download. The bottom-layer EMUs were extracted to assess the distribution of BenthicNet image samples across global benthic environmental regions. Each image was assigned the nearest bottom-layer (i.e. seafloor) EMU in space to compare the sampled frequency of each environment to the proportion of area covered by each EMU (\autoref{fig:image_emu_eg}; \autoref{fig:emus}). A similar analysis was conducted to assess the representativeness of image samples across the broader global oceans by comparing the sampled frequency to the area of each ocean basin, according the the EMU attributes (\autoref{fig:oceans}). The nearest EMU to each image is provided as a metadata field for both the BenthicNet-1M and BenthicNet-Labelled datasets; depths from the GEBCO2022 grid are also provided for each image (assigned using bilinear interpolation; \autoref{tab:dataformat}, \autoref{tab:dataformat_labelled}).

Generally, images were distributed more evenly across the bottom-layer EMUs than would be expected from a random sample, while the distribution across the major ocean basins more closely matched expectation. The majority of the global seafloor (82.4\%) is classified into EMUs 14 (deep, very cold, normal salinity, moderate oxygen, high nitrate, low phosphate, high silicate), 13 (deep, very cold, normal salinity, low oxygen, high nitrate, medium phosphate, high silicate), and 36 (deep, very cold, normal salinity, moderate oxygen, medium nitrate, low phosphate, low silicate) \citep{sayer2017}, comprising most of the Pacific, Indian, and polar oceans. These environments are not over-represented in the BenthicNet dataset, with no single EMU accounting for $>\,20.6\%$. The three most common EMUs sampled (47.6\%) were 24 (shallow, warm, normal salinity, moderate oxygen, low nitrate, low phosphate, low silicate), 11 (shallow, cool, normal salinity, moderate oxygen, low nitrate, low phosphate, low silicate), and 13 (deep, very cold, normal salinity, low oxygen, high nitrate, medium phosphate, high silicate), representing continental shelves in the equatorial regions, the shallow sub-tropics, and the deep Pacific and Indian oceans. The distribution of images across ocean basins was generally proportionate to the expectation given the area of each ocean, but notable exceptions include an apparent under-representation of the South Atlantic, and over-representation of the South Pacific.

\subsubsection{Self-supervised learning}
As a minority of the imagery was labelled, we sought to utilize the unlabelled data by using self-supervised learning (SSL) to train an encoder that may be adapted to downstream tasks on the labelled data.
As a series of benchmarks, we examined four recent SSL methods using a ResNet-50 model architecture, trained on the BenthicNet-1M data. These four methods are from a family of techniques known as instance learning and consist of SimSiam \citep{SimSiam}, Bootstrap Your Own Latent (BYOL) \citep{BYOL}, Momentum Contrast (MoCo-v2) \citep{MoCo-v2}, and Barlow Twins (BT) \citep{BarlowTwins}. We found that overall, the methods performed similarly, with BT performing consistently well at the downstream classification tasks. All subsequent analyses following initial experimentation and reporting uses Barlow Twins as the representative method for SSL.

As an instance learning method, BT's pretext task for learning a useful embedding space works with the encoded representations of a batch of images. Each image in the batch, $X$, is distorted twice using transformations independently randomly selected from a predefined transformation-generator, producing two input views, $X_A$ and $X_B$. The transformation-generator is constructed such that it does not alter the apparent identity of the contents of the image, but does alter other aspects of the image such as the colour balance, contrast, and zoom. Each batch of transformed images is passed through the model to yield embeddings $Z_A$ and $Z_B$. By using the cosine similarity distance metric, a correlation matrix $C$ is constructed between each embedding vector in $Z_A$ and each in $Z_B$. An ideal encoder would be robust against the randomly-selected transforms, producing the same embedding vector no matter which transform is selected, hence we would like the diagonal of $C$ to be 1. Furthermore, images which have different contents should be encoded differently so we can tell them apart from their embeddings; hence we would like the off-diagonal elements of $C$ to be zero. This objective at a high level can be described as rendering the model embedding space invariant to the transformations applied, while also introducing an orthogonality to this model's embedding space \citep{BarlowTwins}.

Using the BT SSL paradigm, we trained a ResNet-50 model on the BenthicNet-1M dataset for three different durations (100, 200, and 400 epochs) with the LARS optimizer \citep{lars}. The hyperparameters were the same as used by \citet{BarlowTwins}, except the learning rate as our initial experiments showed a smaller learning rate of $2 \times 10^{-3}$ yielded better performance than the default of $0.2$. The learning rate was annealed using a one-cycle cosine schedule with a warm-up period of 10 epochs \citep{one-cycle}. The models were trained using four Nvidia A100 GPUs, with a total batch size of 512. The utility of the self-supervised model was evaluated for transfer learning using two tests with labelled data in \textit{Supervised transfer learning}, \autoref{s:Supervised-transfer-learning}, below.

\subsection{Labelled data exploratory analysis}
\begin{figure}[tbh]
    \centerline{
        \includegraphics[scale=1.0]{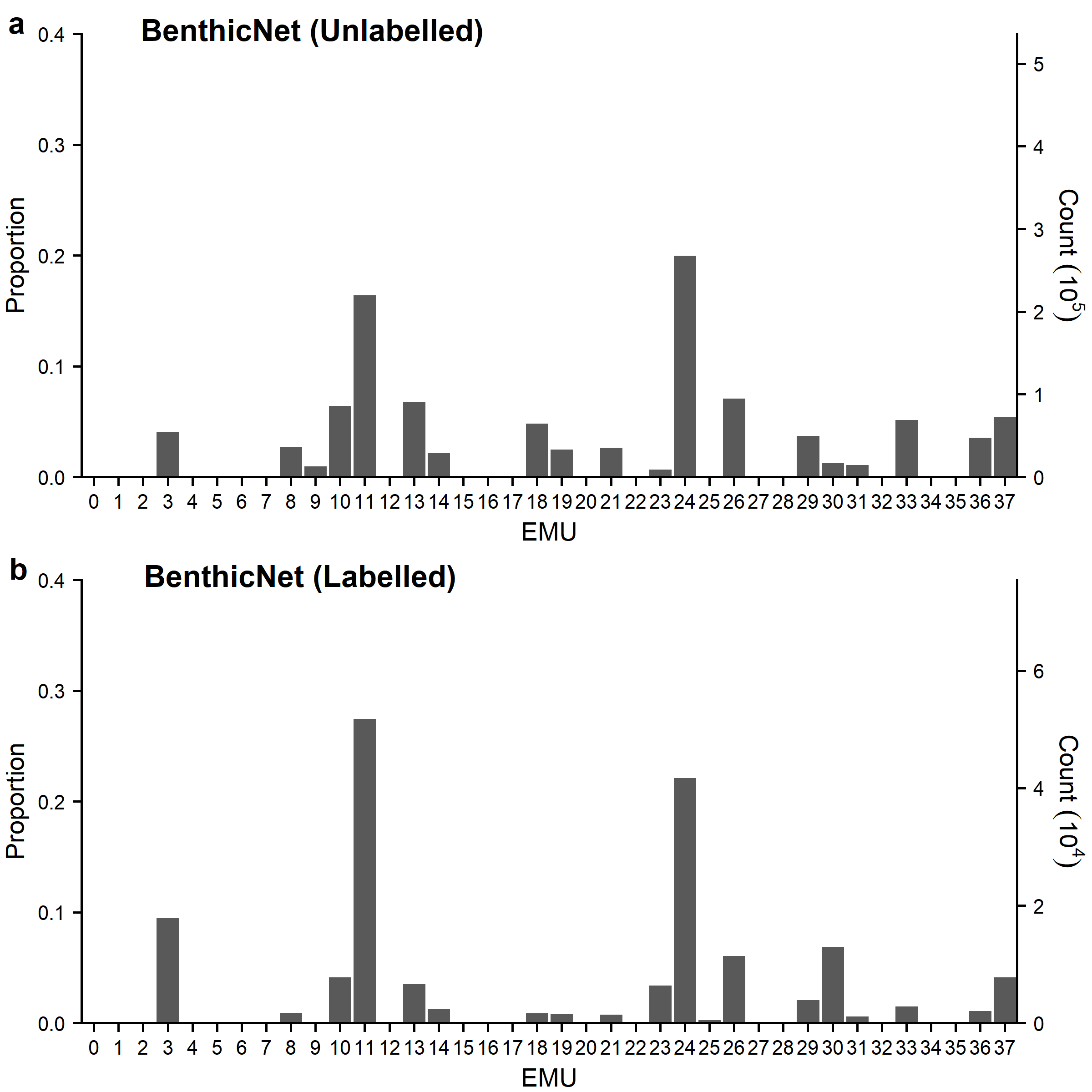}
    }
    \caption{\textbf{Distribution of BenthicNet images according to bottom layer Ecological Marine Units} (EMUs) for (a) unlabelled and (b) labelled datasets. See \citet{sayer2017} for a full description of the EMU classes.}
    \label{fig:emus_labelled}
\end{figure}

The BenthicNet-Labelled data spans an environmental extent similar to that of the BenthicNet-1M data. Two of the EMUs that were abundantly sampled with unlabelled imagery were also prominently represented in the labelled dataset; EMUs 11 (shallow, cool, normal salinity, moderate oxygen, low nitrate, low phosphate, low silicate) and 24 (shallow, warm, normal salinity, moderate oxygen, low nitrate, low phosphate, low silicate) comprised a near-majority (49.82\%) of of the labelled dataset (\autoref{fig:emus_labelled}). These two environments are broadly distributed in space \citep{sayer2017}, and here primarily represent datasets from Australia, Tasmania, and Central America. The full distribution of labels across the CATAMI hierarchy is provided within the dataset hosted on FRDR \citep{BenthicNet_FRDR}.

\subsubsection{Supervised transfer learning}
\label{s:Supervised-transfer-learning}

Here we provide two examples of utilizing a large model pretrained with SSL on the unlabelled BenthicNet-1M dataset for automating benthic image labelling tasks.

First, we trained a model to classify the substrate visible in benthic imagery at the granularity of the second level in the CATAMI substrate hierarchy, comprised of the 5 classes ``Sand/mud'', ``Pebble/gravel'', ``Cobbles'', ``Boulders'', and ``Rock''. This model was trained using the subset of BenthicNet-Labelled containing singly-annotated images with substrate labels at this level. We refer to this subset as the ``BenthicNet-Substrate-d2'' dataset, comprised of \num{57149} images --- \num{43430} of which were used for training, and \num{13719} for testing (partitioned as described above in \textit{Labelled data}, \autoref{s:Labelled-data}). Using the pretrained ResNet-50 backbone, we added a linear classifier head with softmax activation to predict the class of the image. The targets were one-hot encoded. To evaluate the utility of SSL pretraining on the BenthicNet-1M dataset, we also compare against transfer learning from a publicly available ResNet-50 model\maybefootnote{%
\texttt{torchvision.models.ResNet50\_Weights.IMAGENET1K\_V2}\arxivonly{ \href{https://github.com/pytorch/vision/issues/3995\#issuecomment-1013906621}{[recipe]}}%
} pretrained with cross-entropy on ImageNet-1k (600 epochs), provided by torchvision \citep{pytorch}, and against training from scratch on BenthicNet-Substrate-d2 without any pretraining.

Our supervised classification pipeline consists of two stages: a linear probe and fine-tuning. During the linear probe, the pretrained encoder weights are frozen whilst the new linear classifier head is trained. We trained the classifier head for 100 epochs with a one-cycle cosine annealing scheduler for the learning rate. As with the SSL training, the learning rate started at $3 \times 10^{-6}$, and linearly increased to a maximum of $3 \times 10^{-5}$ over 10 warm-up epochs, then was cosine annealed back down to the original rate. For the fine-tuning stage, we begin with the pretrained encoder and the classifier head from the linear probe. We unfreeze the encoder, and train the whole network end-to-end with one tenth the learning rate used for the linear probe for 300 epochs (a total training period of 400 epochs across both stages). 

The transfer-learning models were compared to models trained from scratch. These were randomly initialized, and the entire model (encoder and classifier) trained end-to-end for 100 or 400 epochs, using the one-cycle schedule with peak learning rate $3 \times 10^{-5}$.

\begin{table}[tb]
\centering
\caption{\textbf{Micro-accuracy and macro F1-score (\%) on BenthicNet-Substrate-d2 test data} when training from scratch (No pretraining), with linear probe of a pretrained encoder (frozen, \frozen) or fine tuning (not frozen, \unfrozen). Mean ($\pm$ std. err.) over 3 random seeds (same pretrained backbones over seeds). Bold: best performing linear probe and fine-tuned models.}
\label{tab:sub-2_metrics}
\begin{tabular}{llcrrr}
\toprule
\multicolumn{2}{c}{Pretraining} \\
\cmidrule(lr){1-2}
Dataset       & Loss          & Frozen     & Epochs &  Accuracy $\uparrow$     & F1-score $\uparrow$ \\
\midrule
ImageNet-1k   & Cross-entropy & \frozen    & 100    &         81.8\terr{0.1}  &          56.6\terr{0.3} \\
BenthicNet-1M & Barlow Twins  & \frozen    & 100    & \textbf{83.6}\terr{0.1} &  \textbf{57.7}\terr{0.3} \\
\addlinespace
No pretraining&               & \unfrozen  & 100    &        {81.0}\terr{0.6} &          55.3\terr{1.5} \\
No pretraining&               & \unfrozen  & 400    &         83.8\terr{0.1}  &          61.8\terr{0.3} \\
\addlinespace
ImageNet-1k   & Cross-entropy & \unfrozen  & 100+300& \textbf{88.3}\terr{0.1}  & \textbf{69.5}\terr{0.3} \\
BenthicNet-1M & Barlow Twins  & \unfrozen  & 100+300&        {88.1}\terr{0.1} &          68.5\terr{0.2} \\
\bottomrule
\end{tabular}
\end{table}

\begin{figure}[tb]
    \centerline{
        \includegraphics[scale=1.0]{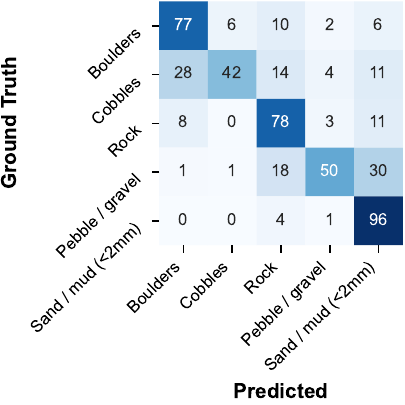}
        \quad
        \includegraphics[scale=1.0]{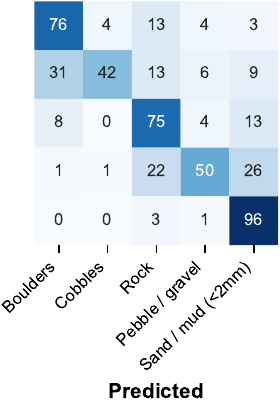}
    }
    \caption{\textbf{Confusion matrix (\% of ground truth) for CATAMI Substrate predictions on BenthicNet-Substrate-d2 test data.} (Left) Model pretrained with cross-entropy on ImageNet-1k, fine-tuned on BenthicNet-Substrate-d2. (Right) Model pretrained with Barlow Twins on BenthicNet-1M, fine-tuned on BenthicNet-Substrate-d2.}
    \label{fig:sub2_confusion}
\end{figure}

As shown in \autoref{tab:sub-2_metrics}, the performance of the fine-tuned ImageNet-1k and BenthicNet-1M pretrained models was comparable when evaluated on unseen BenthicNet-Substrate-d2 test data according to the accuracy and F1-score (the harmonic mean of precision and recall). The fine-tuned (unfrozen) pre-trained models outperformed models trained from scratch, or trained with a frozen pre-trained backbone (a "linear probe"). The linear probe models failed to outperform models trained from scratch for 400 epochs (i.e. with no pre-training). The confusion matrices (\autoref{fig:sub2_confusion}) suggest that the models have similar biases, confusing the same classes as each other (Boulders → Cobbles; Rock → Pebble/gravel; Pebble/gravel → Sand/mud).

As a second task, we considered the German Bank 2010 dataset provided by DFO (\autoref{tab:datasets}), which had whole-image ``benthoscape'' labels described by \citet{german2010}. In the original labelling scheme, five benthoscape labels were assigned that describe recognizable combinations of substrate, bedforms, and biology visible in \num{3181} images, collected off the southwest coast of Nova Scotia, Canada. The benthoscape labels were (1) ``reef'' in which boulders or bedrock with frequent epifauna comprise more than 50\% of images; (2) ``glacial till'' consisting of mixed sediments (cobble, gravel, sand); (3) ``silt/mud'' with frequent evidence of infaunal bioturbation; (4) ``silt with bedforms''; and (5) ``sand with bedforms'', which commonly included sand dollars (\textit{Echinarachnius parma}). Again, the pretrained ResNet-50 models were utilized by adding a new classifier head with outputs corresponding to each of the benthoscape classes. Using the BenthicNet-Labelled partitions (described in \textit{Labelled data}, \autoref{s:Labelled-data}), \num{2681} images were used to train the model and \num{500} were used for testing. The model training hyperparameters were identical to those used for the BenthicNet-Substrate-d2 experiments.

\begin{table}[tbh]
\centering
\caption{\textbf{Micro-accuracy and macro F1-score (\%) on German Bank 2010 test data} when training from scratch (No pretraining), with linear probe of a pretrained encoder (frozen, \frozen) or fine tuning (not frozen, \unfrozen). Mean ($\pm$ std. err.) over $n=3$ random seeds (same pretrained backbones over seeds). Bold: best performing linear probe and fine-tuned models.}
\label{tab:gb2010_metrics}
\begin{tabular}{llcrrr}
\toprule
\multicolumn{2}{c}{Pretraining} \\
\cmidrule(lr){1-2}
Dataset       & Loss          & Frozen     & Epochs &  Accuracy $\uparrow$     & F1-score $\uparrow$ \\
\midrule
ImageNet-1k   & Cross-entropy & \frozen    & 100    &         37.6\terr{5.2}   &         30.0\terr{2.9} \\
BenthicNet-1M & Barlow Twins  & \frozen    & 100    & \textbf{55.9}\terr{2.4}  & \textbf{43.2}\terr{6.0} \\
\addlinespace
No pretraining&               & \unfrozen  & 100    &         53.4\terr{2.4}   &         43.0\terr{3.8} \\
No pretraining&               & \unfrozen  & 400    &         54.1\terr{3.2}   &         46.7\terr{3.2} \\
\addlinespace
ImageNet-1k   & Cross-entropy & \unfrozen  & 100+300&         65.9\terr{4.0}   &         59.2\terr{4.2} \\
BenthicNet-1M & Barlow Twins  & \unfrozen  & 100+300& \textbf{77.0}\terr{0.7}  & \textbf{72.3}\terr{0.8} \\
\bottomrule
\end{tabular}
\end{table}

\begin{figure}[tbh]
    \centerline{
        \includegraphics[scale=1.0]{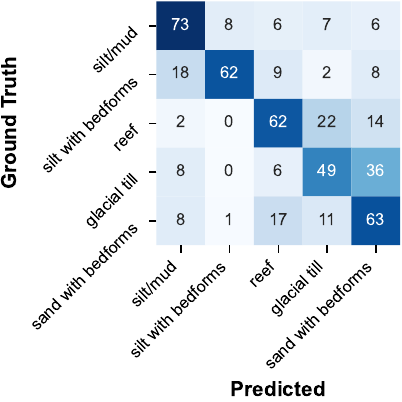}
        \quad
        \includegraphics[scale=1.0]{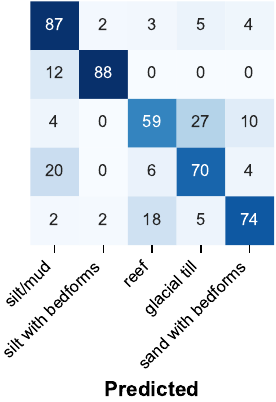}
    }
    \caption{\textbf{Confusion matrix (\% of ground truth) for the German Bank 2010 test data.} (Left) Model pretrained with cross-entropy on ImageNet-1k, fine-tuned on German Bank 2010. (Right) Model pretrained with Barlow Twins on BenthicNet-1M, fine-tuned on German Bank 2010.}
    \label{fig:gb2010_confusion}
\end{figure}

As shown in \autoref{tab:gb2010_metrics}, we observed that the model pretrained on BenthicNet-1M strongly outperformed both the model pretrained on ImageNet-1k and the model trained from scratch when evaluated on unseen test data. The fine-tuned BenthicNet-1M model was able to correctly identify ``silt/mud'' and ``silt with bedforms'' classes in 87\% and 88\% of cases (see \autoref{fig:gb2010_confusion}). 
Both fine-tuned models confused certain class pairs (reef → glacial till; sand with bedforms → reef; silt with bedforms → silt/mud), but the BenthicNet-1M SSL pretrained model was able to greatly increase the recall of both ``silt/mud'' and ``silt with bedforms'', and greatly reduce confusion between other pairs misclassified by the ImageNet-1k model (e.g. glacial till ↔ sand with bedforms).

An important observation is that for both supervised classification tasks, and both transfer models, the best-predicted classes tended to be those that are most distinct, while the intermediate classes were subject to confusion. For example, ``cobble'' was the most difficult label to predict in the BenthicNet-Substrate-d2 dataset, and indeed, it can be difficult even for a human to differentiate cobbles from pebbles or boulders in underwater imagery. These substrate class boundaries are defined arbitrarily at a particular length scale (\SI{2}{\milli\metre} and \SI{64}{\milli\metre}) that may only be determined through accurate measurement or image scaling; there is substantial possibility of incorrect or subjective human labels for such data. Additionally, the imbalanced priors for the classes may also play a role in predictive success. Sand and mud labels dominate both data subsets --- it is not surprising that the models have a tendency to predict sand for other classes, and to perform strongly on sand.

While performance between the transfer models (ImageNet-1k and BenthicNet-1M) were similar for the BenthicNet-Substrate-d2 task, differences in performance on German Bank 2010 were far more pronounced. Notably, our self-supervised model was the strongest performing across all aspects of the evaluation --- outperforming all others using both linear probes and fine tuning \autoref{tab:gb2010_metrics}. The fine-tuned results for BenthicNet-1M also demonstrate less variance compared to the other models. We conjecture that, while transfer learning may perform well for both pretrained models if the labelled dataset is large enough (i.e. tens of thousands of labelled images for the BenthicNet-Substrate-d2 task), the BenthicNet-1M SSL pretrained model is better able to transfer to smaller, more specific classification tasks, where fewer training examples per class are available. Previous exposure to over a million relevant images during the SSL phase may have enhanced the ability to separate images of broadly similar seafloor type. We note that the size of the German Bank 2010 dataset is similar to what might be commonly encountered for a site-specific habitat mapping application (e.g. thousands of labelled images). A number of experiments exploring these and related research questions are currently underway --- for example, on the hierarchical and multi-label CATAMI classification of BenthicNet images \citep{Xu2024a}. Our ResNet-50 model, pretrained on BenthicNet-1M with BT, is accessible from the FRDR repository \citep{BenthicNet_FRDR}. 


\FloatBarrier

\section{Usage Notes}

We note that data labels translated to the CATAMI scheme were sourced from a wide variety of scientific studies with the express intent of supporting the training and validation of large image recognition models. Jointly, these labels should be analyzed with care, particularly if utilized for other purposes. Some datasets included whole image labels indicating the presence of a single benthic feature (e.g. organism, substrate), while others supplied single labels indicating multiple features, or multiple labels for different features within an image. One result of such diversity is variation among the completeness of labels from different datasets --- some, for example, focus on a the presence of single species, or only focus on the most conspicuous or abundant substrate types. For some datasets, it is thus reasonable to expect a larger proportion of false negative labels if the data is treated in a presence/absence manner. In other words, many benthic features are likely visible in the images, which have not been labelled. We operate under the assumption, though, that labels within a dataset were assigned consistently. If performing analyses at the dataset level using the compilation presented here, it is important to investigate the specifics of the dataset(s) in question.

Similarly, the diversity of labelling methodologies has resulted in a number of different schema by which original labels were translated to their CATAMI equivalents. For example, some labels indicating the percent cover of organisms or substrate types in an image were converted to binary presence/absence information for the purposes of assigning labels. Additionally, auxiliary information provided with labels such as annotator notes were used in some cases to obtain a CATAMI label, or to enhance its accuracy. Efforts were made to indicate the original data label as closely as possible in the labelled metadata file, but it was not always possible to include all information that was used to translate an original label to the CATAMI scheme. Therefore, original labels provided in our metadata may not contain all available labelled information for each image, and the original datasets should be referenced as the authoritative source in all cases.

The examples provided here focus on the physical environment, but there are abundant opportunities to explore use of the biological labels. Through use of the SSL pretrained encoder, we believe that the training and deployment of hierarchical morphological and biological identification models is possible. A challenging component of this task is the imbalance of biota labels within the dataset. Methods such as over-sampling, weighting, and data augmentation may be necessary to achieve and validate effective large-scale supervised models in the biota hierarchy of the CATAMI scheme, both to address the label imbalance and the distributional shift from the labelled subset of the data to the full range of ocean imagery. These applications will be explored in coming work.

\section{Code Availability}
\label{s:Code-availability}

Code used to query, download, convert, process, subsample, and partition the full data compilation may be accessed without restriction from \url{https://github.com/DalhousieAI/BenthicNet}. Code used to query and download data from SQUIDLE+ using the API is available at \url{https://github.com/DalhousieAI/squidle-downloader}. Code used to query and download data from PANGAEA is available at \url{https://github.com/DalhousieAI/pangaea-downloader}.

Code used to train the self-supervised model is available at \url{https://github.com/DalhousieAI/ssl-bentho}. Code used to perform one-hot multi-class transfer learning, as presented in \textit{Supervised transfer learning}, \autoref{s:Supervised-transfer-learning}, is available at \url{https://github.com/DalhousieAI/benthicnet_probes}.

\section*{Acknowledgements}

This research is part of the Ocean Frontier Institute (\href{https://www.ofi.ca/}{OFI}) Benthic Ecosystem Mapping and Engagement (\href{https://www.ofibecome.org/}{BEcoME}) Project.

Data was sourced from Australia’s Integrated Marine Observing System (\href{https://imos.org.au/}{IMOS}) --- IMOS is enabled by the National Collaborative Research Infrastructure Strategy (\href{https://www.education.gov.au/ncris}{NCRIS}). It is operated by a consortium of institutions as an unincorporated joint venture, with the University of Tasmania as Lead Agent.

We would would like to acknowledge the Australian National Research Program and the Australian Centre for Field Robotics (\href{https://www.sydney.edu.au/engineering/our-research/robotics-and-intelligent-systems/australian-centre-for-robotics.html}{ACFR}) for gathering image data that was used as part of this project.

We are grateful to the many Reef Life Survey (\href{https://reeflifesurvey.com/}{RLS}) divers around the world who contributed to data that was sourced for this project.

NGU data collection was funded partly by \href{https://www.ngu.no/en}{NGU} and partly by local/regional communities in various Marine Base Maps projects between 2010 and 2017.

A large number of images sourced from \href{https://squidle.org/}{SQUIDLE+} were collected as part of the Reef Builder program 2021--2023, led by The Nature Conservancy Australia and supported by the Australian Government.

For the data from western Canada, we thank Hakai Institute staff Nick Viner for collecting and preparing ROV footage and Keith Holmes, Derek VanMaanan, Carolyn Knapper, Ben Millard-Martin and Ondine Pontier for support with towed video collection and analysis. We also thank colleagues at the Pacific Rim National Park Reserve --- Jennifer Yakimishyn, Caron Olive, Mike Collyer, Angela Rehhorn, and Silvana Botros --- for their support in the collection and analysis of towed video for seagrass mapping. The data contributed by the Hakai Institute was collected within the traditional territories of the Haí\textltilde zaqv (Heiltsuk) and Wuikinuxkv First Nations on the Central Coast of British Columbia and the Tseshaht and nuučaa\'nuu\textltilde \textglotstop atḥ nis\'ma (Nuu-chah-nulth) Nations on the west coast of Vancouver Island.

Thanks to Yan Liang Tan and Molly Wells for contributions to CATAMI label translation.

Thank you to Erin Clary for patient and detailed curation of BenthicNet, and for facilitating data hosting with the Federated Research Data Repository. 

This research was enabled in part by support provided by \href{https://ace-net.ca/}{ACENET} and the Digital Research Alliance of Canada (\href{https://alliancecan.ca/}{alliancecan.ca}).

This research was enabled in part by support provided by the \href{https://deepsense.ca/}{DeepSense} computing platform. DeepSense is funded by the Atlantic Canada Opportunities Agency (\href{https://www.canada.ca/en/atlantic-canada-opportunities.html}{ACOA}), the Province of Nova Scotia, the Centre for Ocean Ventures and Entrepreneurship (\href{https://coveocean.com/}{COVE}), IBM Canada~Ltd., and the Ocean Frontier Institute (\href{https://www.ofi.ca/}{OFI}).

\section*{Author Contributions}

S.C.L. and B.M. conceived of and designed the project. S.C.L., B.M., and A.B. acquired publicly available data from online repositories. S.C.L., B.M., and I.X. wrote the initial manuscript draft. S.C.L. and B.M. designed the data partitioning scheme. S.C.L. and I.X. designed and executed the data sub-sampling routine. S.C.L., B.M., I.X., S.A., and K.M. contributed to CATAMI label translation. I.X. conducted the modelling. C.J.B. and T.T. supervised the project; and K.R., C.J.B., and T.T. acquired funding to support it. S.C.L., B.M., A.C.B., M.B., V.F., A.F., D.H., D.I., J. M.-M., K.M., P.S.M., J.M., S.N., J.O., E.O., L.Y.R., K.R., C.M.R., J.A.S., A.C.G.S., J.A.T., B.R.W., M.C.W., C.J.B. contributed data that was used to establish the BenthicNet data compilation and models. S.C.L., B.M., I.X., A.C.B., M.B., V.F., A.F., D.H., D.I., J.M.-M., K.M., P.S.M., J.M., S.N., J.O., E.O., L.Y.R., K.R., C.M.R., J.A.S., A.C.G.S., J.A.T., B.R.W., M.C.W., C.J.B., T.T. contributed to editing and revision of the initial draft. All authors reviewed the manuscript.

\section*{Competing Interests}

The authors declare no competing interests.

\bibliographystyle{apalike-url-x}
\bibliography{references, pangaea-datasets_2024-02-08_ben, pangaea-refs_ben, noaa-refs, usgs-refs, usap-dc-refs, aadc-refs, mgds-refs}

\clearpage
\appendix
\section*{\LARGE Appendices}

\section{PANGAEA Search}
\label{a:pangaea-search}

To thoroughly search PANGAEA for seafloor imagery, we used 20 search terms with a range of synonyms for the content of interest.
The PANGAEA search API is comprehensive and allows terms be combined with AND or OR operators, and negative search terms to be used.
However, we could not merge all our synonyms together into a single, large query because the number of results which can be returned by one query is limited to 500 records.

The search terms used were as follows:
{\small
\begin{verbatim}
(seabed OR "sea bed" OR "sea-bed") (image OR imagery OR photo OR photograph
    OR "photo-transect" OR photoquad* OR photo-quad* OR jpg OR jpeg OR png OR tif
    OR tiff)
(seafloor OR "sea floor" OR "sea-floor") (image OR imagery OR photo OR photograph
    OR "photo-transect" OR photoquad* OR photo-quad* OR jpg OR jpeg OR png OR tif
    OR tiff)
("ocean floor" OR "ocean-floor") (image OR imagery OR photo OR photograph
    OR "photo-transect" OR photoquad* OR photo-quad* OR jpg OR jpeg OR png OR tif
    OR tiff)
underwater (habitat* OR substrate OR sediment) (image OR imagery OR photo
    OR photograph OR "photo-transect" OR photoquad* OR photo-quad* OR jpg OR jpeg
    OR png OR tif OR tiff)
benthic (image OR imagery OR photo OR photograph OR "photo-transect" OR photoquad*
    OR photo-quad* OR jpg OR jpeg OR png OR tif OR tiff)
(benthos or benthoz) (image OR imagery OR photo OR photograph OR "photo-transect"
    OR photoquad* OR photo-quad* OR jpg OR jpeg OR png OR tif OR tiff)
(coral OR reef OR seagrass OR "sea grass") (image OR imagery OR photo OR photograph
    OR "photo-transect" OR photoquad* OR photo-quad* OR jpg OR jpeg OR png OR tif
    OR tiff)
(auv OR rov OR uuv OR "underwater vehicle") (image OR imagery OR photo OR photograph
    OR "photo-transect" OR photoquad* OR photo-quad* OR jpg OR jpeg OR png OR tif
    OR tiff)
benthoscape habitat* image
benthoscape habitat* imagery
benthoscape habitat* photo
benthoscape habitat* photograph
benthoscape habitat* ("photo-transect" OR photoquad* OR photo-quad*)
benthoscape habitat* (jpg OR jpeg OR png OR tif OR tiff)
benthoscape image
benthoscape imagery
benthoscape photo
benthoscape photograph
benthoscape ("photo-transect" OR photoquad* OR photo-quad*)
benthoscape (jpg OR jpeg OR png OR tif OR tiff)
\end{verbatim}
}

Each search term was prefixed with a set of negative search terms to remove false positives, given as follows
{\small
\begin{verbatim}
-microscop? -"Meteorological observations" -topsoil -soil -sky
    -"wind vector" -"wind stress" -"vertical profile" -"vertical distribution"
\end{verbatim}
}

The full code for our PANGAEA search is publicly available at \\\url{https://github.com/DalhousieAI/pangaea-downloader}.

\newpage
\section{FathomNet Python API Code}
\label{c:fathomnet_code}

We retrieved the full set of images on FathomNet by using the FathomNet API from the \href{https://fathomnet-py.readthedocs.io/en/latest/api.html}{fathonnet-py} Python package as follows.

\noindent\rule{\columnwidth}{0.4pt}
{\small
\begin{Verbatim}[commandchars=\\\{\}]
\PYG{k+kn}{import} \PYG{n+nn}{fathomnet.api.images}
\PYG{k+kn}{import} \PYG{n+nn}{pandas} \PYG{k}{as} \PYG{n+nn}{pd}

\PYG{n}{keys} \PYG{o}{=} \PYG{p}{[}\PYG{l+s+s2}{\PYGZdq{}url\PYGZdq{}}\PYG{p}{,} \PYG{l+s+s2}{\PYGZdq{}uuid\PYGZdq{}}\PYG{p}{,} \PYG{l+s+s2}{\PYGZdq{}timestamp\PYGZdq{}}\PYG{p}{,} \PYG{l+s+s2}{\PYGZdq{}latitude\PYGZdq{}}\PYG{p}{,} \PYG{l+s+s2}{\PYGZdq{}longitude\PYGZdq{}}\PYG{p}{]}

\PYG{n}{records} \PYG{o}{=} \PYG{p}{[]}
\PYG{k}{for} \PYG{n}{submitter} \PYG{o+ow}{in} \PYG{n}{fathomnet}\PYG{o}{.}\PYG{n}{api}\PYG{o}{.}\PYG{n}{images}\PYG{o}{.}\PYG{n}{find\PYGZus{}distinct\PYGZus{}submitter}\PYG{p}{():}
    \PYG{k}{for} \PYG{n}{image} \PYG{o+ow}{in} \PYG{n}{fathomnet}\PYG{o}{.}\PYG{n}{api}\PYG{o}{.}\PYG{n}{images}\PYG{o}{.}\PYG{n}{find\PYGZus{}by\PYGZus{}contributors\PYGZus{}email}\PYG{p}{(}\PYG{n}{submitter}\PYG{p}{):}
        \PYG{n}{records}\PYG{o}{.}\PYG{n}{append}\PYG{p}{(\PYGZob{}}\PYG{n}{k}\PYG{p}{:} \PYG{n+nb}{getattr}\PYG{p}{(}\PYG{n}{image}\PYG{p}{,} \PYG{n}{k}\PYG{p}{)} \PYG{k}{for} \PYG{n}{k} \PYG{o+ow}{in} \PYG{n}{keys}\PYG{p}{\PYGZcb{})}

\PYG{n}{df} \PYG{o}{=} \PYG{n}{pd}\PYG{o}{.}\PYG{n}{DataFrame}\PYG{o}{.}\PYG{n}{from\PYGZus{}records}\PYG{p}{(}\PYG{n}{records}\PYG{p}{)}
\PYG{n}{df}\PYG{o}{.}\PYG{n}{drop\PYGZus{}duplicates}\PYG{p}{(}\PYG{n}{subset}\PYG{o}{=}\PYG{l+s+s2}{\PYGZdq{}url\PYGZdq{}}\PYG{p}{,} \PYG{n}{inplace}\PYG{o}{=}\PYG{k+kc}{True}\PYG{p}{)}
\end{Verbatim}
}
\noindent\rule{\columnwidth}{0.4pt}

\end{document}